%% file: output_arvix.tex
\documentclass[10pt,twocolumn,letterpaper]{article}

\usepackage[pagenumbers]{cvpr} 

\input{preamble}

\definecolor{cvprblue}{rgb}{0.21,0.49,0.74}
\usepackage[pagebackref,breaklinks,colorlinks,citecolor=cvprblue]{hyperref}
\usepackage{array}

\usepackage{graphicx}
\makeatletter
\newcommand*\bigcdot{\mathpalette\bigcdot@{.5}}
\newcommand*\bigcdot@[2]{\mathbin{\vcenter{\hbox{\scalebox{#2}{$\m@th#1\bullet$}}}}}

\def\paperID{16406} 
\def\confName{CVPR}
\def\confYear{2024}

\title{Boosting Latent Diffusion with Flow Matching}

\author{
Johannes Schusterbauer$^*$
\qquad
Ming Gui$^*$
\qquad 
Pingchuan Ma$^*$ \\
Nick Stracke
\qquad
Stefan A. Baumann 
\qquad
Vincent Tao Hu
\qquad 
Björn Ommer
\\
\\
CompVis @ LMU Munich, MCML
}

\begin{document}

\input{figures/LCM_v15_xl_comparison}

\input{sec/0_Abstract}


\section{Introduction}
\label{sec:intro}
\input{sec/1_Intro}

\section{Related Work}
\label{sec:relatedwork}
\input{sec/2_RelatedWork}

\section{Method}
\label{sec:method}
\input{sec/3_Method}

\section{Experiments}
\label{sec:experiments}
\input{sec/4_Experiments}

\section{Conclusion}
\label{sec:conclusion}
\input{sec/5_Conclusion}

{
    \small
    \bibliographystyle{ieeenat_fullname.bst}
    \bibliography{ref.bib}
}

\clearpage
\input{sec/6_Appendix}

\end{document}

%% file: preamble.tex
\usepackage[dvipsnames]{xcolor}

\newcommand{\da}{\downarrow}
\newcommand{\ua}{\uparrow}
\newcommand{\und}[1]{\underline{#1}}

\usepackage{comment}
\usepackage{multirow}
\usepackage{booktabs}
\usepackage{url}
\usepackage{epsfig}
\usepackage{graphicx}
\usepackage{amsmath}
\usepackage{amssymb}
\usepackage{subcaption} 
\usepackage{enumitem}
\usepackage{wrapfig}

\usepackage{multibib}

\usepackage{numprint}
\usepackage[figuresleft]{rotating}
\usepackage{float}
\usepackage{caption}
\usepackage{adjustbox}

\usepackage{bbding}
\usepackage{tikz}
\usetikzlibrary{angles,quotes,3d,math,arrows.meta,calc,positioning,fit,backgrounds,decorations.pathreplacing,calligraphy,shapes,shapes.multipart}

%% file: figures/LCM_v15_xl_comparison.tex
\twocolumn[{%
		\renewcommand\twocolumn[1][]{#1}%
		\maketitle
            \vspace{-2em}
		\begin{center}
			\includegraphics[width=.95\textwidth]{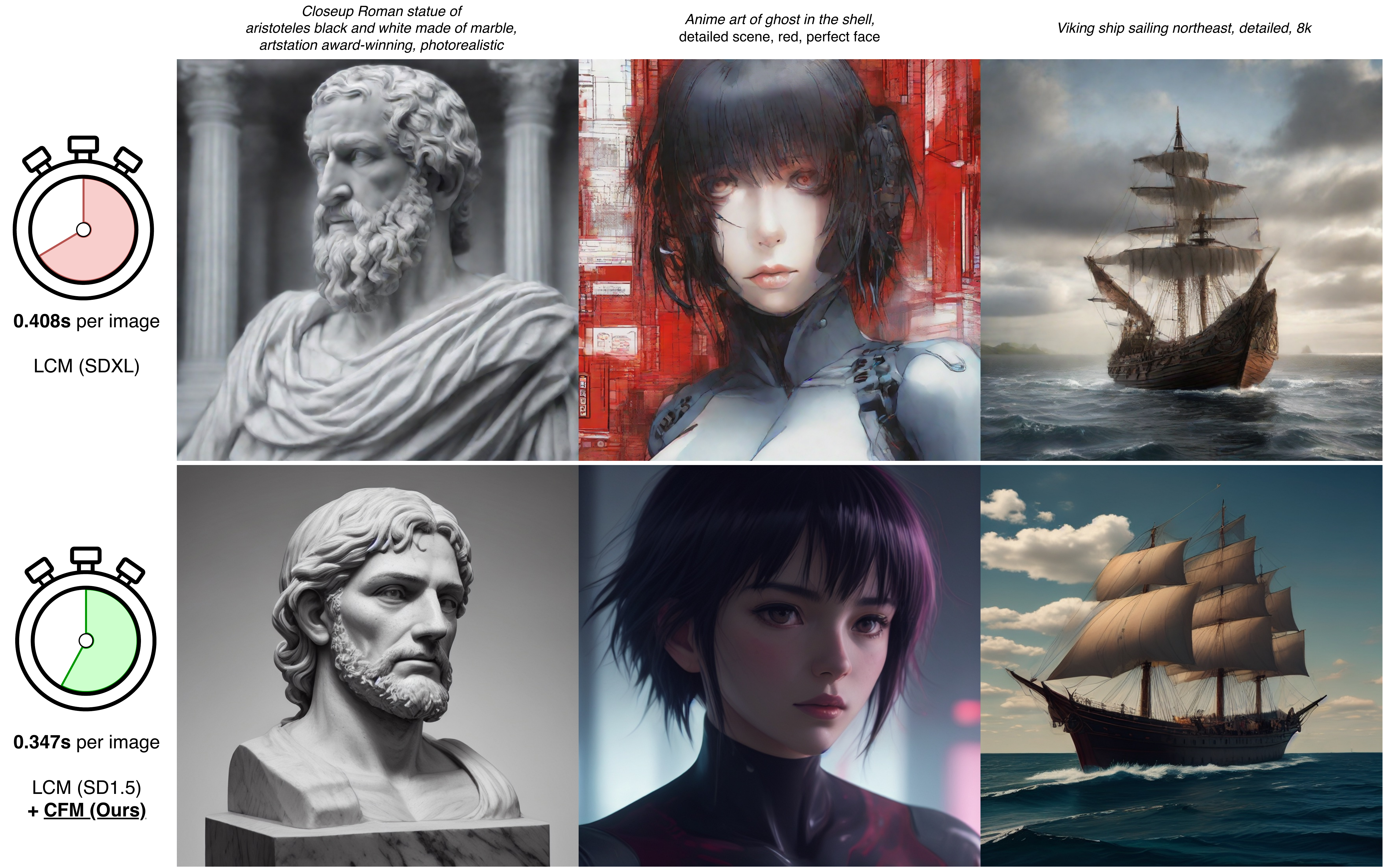}
			\captionsetup{type=figure}
			\captionof{figure}{
				\textbf{Samples synthesized in $1024^2$ px.} We elevate Diffusion Models (DMs) and similar architectures to a higher-resolution domain, achieving exceptionally rapid processing speeds. We use Latent Consistency Models (LCM) \cite{luo2023lcm}, distilled from SD1.5 \cite{rombach2022high_latentdiffusion_ldm} and SDXL \cite{podell2023sdxl}, respectively. To achieve the same resolution as LCM-SDXL, we boost LCM-SD1.5 with our Coupling Flow Matching (CFM) model. The LCM-SDXL model fails to produce competitive results within this shortened timeframe, highlighting the effectiveness of our approach in achieving both speed and quality in image synthesis.
    }
    \label{fig:LCM_SD1.5vsXL}
		\end{center}
	}]

 \def\thefootnote{*}\footnotetext{Equal Contribution}

%% file: sec/0_Abstract.tex
\begin{abstract}
Visual synthesis has recently seen significant leaps in performance, largely due to breakthroughs in generative models. Diffusion models have been a key enabler, as they excel in image diversity. However, this comes at the cost of slow training and synthesis, which is only partially alleviated by latent diffusion. To this end, flow matching is an appealing approach due to its complementary characteristics of faster training and inference but less diverse synthesis.
We demonstrate that introducing flow matching between a frozen diffusion model and a convolutional decoder enables high-resolution image synthesis at reduced computational cost and model size. A small diffusion model can then effectively provide the necessary visual diversity, while flow matching efficiently enhances resolution and detail by mapping the small to a high-dimensional latent space. These latents are then projected to high-resolution images by the subsequent convolutional decoder of the latent diffusion approach. Combining the diversity of diffusion models, the efficiency of flow matching, and the effectiveness of convolutional decoders, state-of-the-art high-resolution image synthesis is achieved at $1024^2$ pixels with minimal computational cost. Further scaling up our method we can reach resolutions up to $2048^2$ pixels. Importantly, our approach is orthogonal to recent approximation and speed-up strategies for the underlying model, making it easily integrable into the various diffusion model frameworks. Project page and code are available at \url{https://compvis.github.io/fm-boosting}.
\end{abstract}

%% file: sec/1_Intro.tex
Visual synthesis has recently witnessed unprecedented progress and popularity in computer vision and beyond. Various generative models have been proposed to address the diverse challenges in this field \cite{xiao2021trilemma}, including sample diversity, quality, resolution, training, and test speed. Among these approaches, diffusion models (DMs) \cite{saharia2022imagen, ramesh2022dalle2, rombach2022high_latentdiffusion_ldm} currently rank among the most popular and highest quality, defining the state of the art in numerous synthesis applications. While DMs excel in sample quality and diversity, they face challenges in high-resolution synthesis, slow sampling speed, and a substantial memory footprint.

Lately, numerous efficiency improvements to DMs have been proposed \cite{chen2023pixart,song2020denoising_ddim, rabe2021xformers}, but the most popular remedy has been the introduction of Latent Diffusion Models (LDMs)~\cite{rombach2022high_latentdiffusion_ldm}. Operating only in a compact latent space, LDMs combine the strengths of DMs with the efficiency of a convolutional encoder-decoder that translates the latents back into pixel space. However, Rombach et al.~\cite{rombach2022high_latentdiffusion_ldm} also showed that an excessively strong first-stage compression leads to information loss, limiting generation quality. Efforts have been made to expand the latent space \cite{podell2023sdxl} or stack a series of different DMs, each specializing in different resolutions \cite{saharia2022imagen, ho2022cascaded}. However, these approaches are still computationally costly, especially when synthesizing high-resolution images.

The inherent stochasticity of DMs is key to their proficiency in generating diverse images. In the later stages of DM inference, as the global structure of the image has already been generated, the advantages of stochasticity diminish. Instead, the computational overhead due to the less efficient stochastic diffusion trajectories becomes a burden rather than helping in up-sampling to and improving higher resolution images \cite{balaji2022ediffi}.
At this stage, converse characteristics become beneficial: reduced diversity and a short and straight trajectory toward the high-resolution latent space of the decoder. These goals align precisely with the strengths of Flow Matching (FM) \cite{lipman2023flow,albergo2023stochastic,liu2023rectified}, another emerging family of generative models currently gaining significant attention. In contrast to DMs, Flow Matching enables the modeling of an optimal transport conditional probability path between two distributions that is significantly straighter than those achieved by DMs, making it more robust, and efficient to train. The deterministic nature of Flow Matching models also allows the utilization of off-the-shelf Ordinary Differential Equation (ODE) solvers, which are more efficient to sample from and can further accelerate inference.

We leverage the complementary strengths of DMs, FMs, and VAEs: the diversity of stochastic DMs, the speed of Flow Matching in training and inference stages, and the efficiency of a convolutional decoder when mapping latents into pixel space. This synergy results in a small diffusion model that excels in generating diverse samples at a low resolution. Flow Matching then takes a direct path from this lower-resolution representation to a higher-resolution latent, which is subsequently translated into a high-resolution image by a convolutional decoder. %
Moreover, the Flow Matching model can establish data-dependent couplings with the synthesized information from the DM, which automatically and inherently forms optimal transport paths from the noise to the data samples in the Flow Matching model~\cite{tong2023improving,albergo2023couplings}.

Note that our work is complementary to recent work on sampling acceleration of diffusion models like DDIM~\cite{song2020denoising_ddim}, DPM-Solver~\cite{lu2022dpm++}, and LCM-LoRA \cite{luo2023latent,luo2023lcm}. Our approach can be directly integrated into any existing DMs architecture to increase the final output resolution efficiently.

%% file: sec/2_RelatedWork.tex
\textbf{Diffusion Models}
Diffusion models~\cite{sohl2015deep,ho2020denoising,song2021scorebased_sde} have shown broad applications in computer vision, spanning image~\cite{rombach2022high_latentdiffusion_ldm}, audio~\cite{liu2023audioldm}, and video~\cite{ho2022video,blattmann2023align_videoldm}. Albeit with high fidelity in generation, they do so at the cost of sampling speed compared to alternatives like Generative Adversarial Networks \cite{goodfellow2020gans, kang2023gigagan, karras2019styleGAN}.
Hence, several works propose more efficient sampling techniques for diffusion models, including distillation~\cite{salimans2022progressive,song2023consistency, meng2023distillation}, noise schedule design~\cite{kingma2021variational,nichol2021improved,preechakul2022diffusion_autoencoder}, and training-free sampling~\cite{song2020denoising_ddim,karras2022elucidating,lu2022dpm,liu2022pseudo_pndm}. Nonetheless, it is important to highlight that existing methods have not fully addressed the challenge imposed by the strong curvature in the sampling trajectory~\cite{lee2023minimizing}, which limits sampling step sizes and necessitates the utilization of intricately tuned solvers, making sampling costly.

\noindent\textbf{Flow Matching-based Generative Models}
A recent competitor, known as Flow Matching~\cite{lipman2023flow, liu2023rectified,albergo2022building,neklyudov2023action}, has gained prominence for its ability to maintain straight trajectories during generation by modeling the synthesis process using an optimal transport conditional probability path with Ordinary Differential Equations (ODE), positioning it as an apt alternative for addressing trajectory straightness-related issues encountered in diffusion models. The versatility of Flow Matching has been showcased across various domains, including image~\cite{lipman2023flow, dao2023flowlatent, taohu2023lfm}, video~\cite{video_fm}, audio~\cite{le2023voicebox}, and depth estimation \cite{gui2024depthfm}.
This underscores its capacity to address the inherent trajectory challenges associated with diffusion models, mitigating the limitations of slow sampling in the current generation based on diffusion models. 
Considerable effort has been directed towards optimizing transport within Flow Matching models \cite{tong2023improving, liu2023rectified}, which contributes to enhanced training stability and accelerated inference speed by making the trajectories even straighter and thus enabling larger sampling step sizes.
However, the generation capabilities of Flow Matching presently do not parallel those of diffusion models \cite{lipman2023flow,dao2023flowlatent}. We remedy this limitation by using a small diffusion model for synthesis quality.

\noindent\textbf{Image Super-Resolution}
Image super-resolution (SR) is a fundamental problem in computer vision. Prominent methodologies include GANs \cite{ledig2017srgan, wang2018esrgan, zhang2021bsrgan, kang2023gigagan}, diffusion models \cite{yue2024resshift, saharia2022_SR3, li2022srdiff} and Flow Matching methods \cite{lipman2023flow,albergo2023couplings}. 

Our methodology adopts the Flow Matching approaches, leveraging it's objective to achieve faster training and inference compared to diffusion models. We take inspiration from latent diffusion models \cite{rombach2022high_latentdiffusion_ldm} and transition the training to the latent space, which further enhances computational efficiency. This enables the synthesis of images with significantly higher resolution, thereby advancing the capacity for image generation in terms of both speed and output resolution.

%% file: sec/3_Method.tex
\input{figures/pipeline}
\input{figures/128_to_2k}

We speed up and increase the resolution of existing LDMs by integrating Flow Matching in the latent space.
The proposed architecture should not be limited to unconditional image synthesis but also be applicable to text-to-image synthesis \cite{rombach2022high_latentdiffusion_ldm, nichol2021glide, saharia2022imagen, ramesh2022dalle2} and Diffusion models with other conditioning including depth maps, canny edges, etc.~\cite{esser2023gen1, zhang2023controlnet, li2023gligen}.
The main challenge is not a deficiency in diversity within the Diffusion model; rather it is the slow convergence of the training procedure, the huge memory demand, and the slow inference~\cite{xue2023raphael,podell2023sdxl, saharia2022imagen}. 

While there are substantial efforts to accelerate inference speed of DMs either by distillation techniques \cite{meng2023distillation}, or by an ODE approximation at inference \cite{song2020denoising_ddim,lu2022dpm,lu2022dpm++}, we argue that we can achieve faster training and inference speed by training with an ODE assumption \cite{lipman2023flow}.

Flows characterized by straight paths without Wiener process inherently incur minimal time-discretization errors during numerical simulation~\cite{liu2023rectified} and can be simulated with only few ODE solver steps.

We employ a compact Diffusion model and a Flow Matching model aimed at high-resolution image generation (\cref{sec:diffusion}, \cref{sec:flow_matching}). The combination of both models (\cref{sec:FM+DM}) ensures efficient and detailed image generation.

\subsection{From LDM to FM-LDM}
\label{sec:diffusion}

Diffusion Models (DMs) \cite{ho2020denoising} are generative models that learn a data distribution $p(x)$ by learning to denoise noisy samples. During inference, they generate samples in a multi-step denoising process starting from Gaussian noise.
Their inherent stochasticity allows them to effectively approximate the data manifold with high diversity, even in high-dimensional complex data domains such as images~\cite{saharia2022imagen,nichol2021improved} or videos \cite{ho2022video,singer2022makeavideo,blattmann2023align_videoldm}, but makes generation inefficient, requiring many denoising steps at the data resolution. This problem has previously partially been addressed by \textit{Latent Diffusion Models} (LDMs), which move the diffusion process to an autoencoder latent space, but efficiency is still a problem.
While diffusion models' stochasticity helps them generate high-quality samples, we propose that this stochasticity is not needed for later stages of generation and that the diffusion generation process can be separated into two parts without substantial loss in quality: one diffusion-based low-resolution stage for generating image semantics with high variation and a light-weight high-resolution stage with reduced stochasticity.

Recently, the formulation of generative processes as optimal transport conditional probability paths has gained much attraction \cite{albergo2022building,tong2023improving,lipman2023flow}, perfectly suiting this task of modeling straight trajectories between two distributions.

\subsection{Flow Matching}
\label{sec:flow_matching}
Flow Matching models are generative models that regress vector fields based on fixed conditional probability paths. Let $\mathbb{R}^d$ be the data space with data points $x$. 
Let $u_t(x): [0,1] \times \mathbb{R}^d \rightarrow \mathbb{R}^d$ be the time-dependent vector field, which defines the ODE in the form of $dx=u_t(x)dt$, and let $\phi_t(x)$ denote the solution to this ODE with the initial condition $\phi_0(x)=x$.

The probability density path $p_t:[0,1]\times \mathbb{R}^d \rightarrow \mathbb{R}_{>0}$ depicts the probability distribution of $x$ at timestep $t$ with $\int p_t(x)dx=1$. The pushforward function $p_t=[\phi_t]_{\#}(p_0)$ then transports the probability density path $p$ along $u$ from timestep $0$ to $t$.
Assuming that $p_t(x)$ and $u_t(x)$ are known, and the vector field $u_t(x)$ generates $p_t(x)$, we can regress a vector field $v_\theta(t,x)$ parameterized by a neural network with learnable parameters $\theta$ using the Flow Matching objective
\begin{equation}
    \mathcal{L}_{FM}(\theta)=\mathbb{E}_{t,p_t(x)}||v_\theta(t,x)-u_t(x)||.
\end{equation}
 While we generally do not have access to a closed form of $u_t$ because this objective is intractable, Lipman et al. \cite{lipman2023flow} showed that we can acquire the same gradients and therefore efficiently regress the neural network using the coupling Flow Matching (CFM) objective, where we can compute $u_t(x|z)$ by efficiently sampling $p_t(x|z)$,
\begin{equation}
    \mathcal{L}_{CFM}(\theta)=\mathbb{E}_{t,q(z),p_t(x|z)}||v_\theta(t,x)-u_t(x|z)||,
\end{equation}
\noindent with $z$ as a conditioning variable and $q(z)$ the distribution of that variable.
We parameterize $v_\theta$ as a U-Net \cite{ronneberger2015unet}, which takes the data sample $x$ as input and $z$ as conditioning information.

\subsubsection{Na\"ive Flow Matching}

We first assume that the probability density path starts from $p_0$ with standard Gaussian distribution and ends up in a Gaussian distribution $\mathcal{N}(x_1, \sigma_{\min}^2)$ that is smoothed around a data sample $x_1$ with minimal variance. In this case, the conditioning signal $z$ would be $x_1$, and the optimal transportation path would be formulated as follows \cite{lipman2023flow},
\begin{equation}
    p_t(x|z)=\mathcal{N}(x|tx_1,(t\sigma_{\min} -t+1)^2\mathbf{I}),
\end{equation}
\begin{equation}
    u_t(x|z)=\frac{x_1-(1-\sigma_{\min})x}{1-(1-\sigma_{\min})t};
\end{equation}
\begin{equation}
    \phi_t(x|z)=(1-(1-\sigma_{\min})t)x + tx_1.
\end{equation}
\noindent The resulting FM loss takes the form of
\begin{equation}
    \begin{split} 
        \mathcal{L}_{FM}(\theta) &= \mathbb{E}_{t,z,p_t(x|z)}||v_\theta(t,\phi_t(x_0))- \frac{d}{dt}\phi_t(x_0)|| \\
        & = \mathbb{E}_{t,z,p(x_0)}||v_\theta(t,\phi_t(x_0))- (x_1-(1-\sigma_{\min})x_0)||.
    \end{split}
    \label{eq:loss}
\end{equation}

\subsubsection{Data-Dependent Couplings}

In our case, we also have access to the representation of a low-resolution image generated by a DM at inference time. It seems intuitive to incorporate the inherent relationship between the conditioning signal and our target within the Flow Matching objective, as is also stated in \cite{albergo2023couplings}.
Let $x_1$ denote a high-resolution image. The conditioning signal $z:=x_1$ remains unchanged from the previous formulation.
Instead of randomly sampling from a Gaussian distribution in the na\"ive Flow Matching method, the starting point $x_0 = \mathcal{E}(x_1)$ corresponds to an encoded representation of the image, with $\mathcal{E}$ being a fixed encoder. 

Similar to the previously described case, we smooth around the data samples within a minimal variance to acquire the corresponding data distribution $\mathcal{N}(x_0, \sigma_{\min}^2)$ and $\mathcal{N}(x_1, \sigma_{\min}^2)$. The Gaussian flows can be defined by the equations
\begin{equation}
    p_t(x|z)=\mathcal{N}(x|tx_1+(1-t)x_0,\sigma_{\min}^2\mathbf{I}),
\end{equation}
\begin{equation}
    u_t(x|z)=x_1 - x_0; \ \ \ \phi_t(x|z)=tx_1+(1-t)x_0.
\end{equation}

Notably, the optimal transport condition between the probability distributions $p_0(x|z)$ and $p_1(x|z)$ is inherently satisfied due to the data coupling. This automatically solves the dynamic optimal transport problem in the transition from low to high resolution within the Flow Matching paradigm and enables more stable and faster training \cite{tong2023improving}. We name these Flow Matching models with data-dependent couplings \textit{Coupling Flow Matching} (CFM) models, and the CFM loss then takes the form of
\begin{equation}
    \mathcal{L}_{CFM}(\theta)= \mathbb{E}_{t,z,p(x_0)}||v_\theta(t,\phi_t(x_0))- (x_1-x_0)||.
\end{equation}

\subsubsection{Noise Augmentation}

Noise Augmentation is a technique for boosting generative models' performance introduced for cascaded Diffusion models \cite{ho2022cascaded}. The authors found that applying random Gaussian noise or Gaussian blur to the conditioning signal in super-resolution Diffusion models results in higher-quality results during inference. Drawing inspiration from this, we also implement Gaussian noise augmentation on $x_0$. Following variance-preserving DMs, we noise $x_0$ according to the cosine schedule first proposed in \cite{nichol2021improved}. In line with \cite{ho2022cascaded}, we empirically discover that incorporating a specific amount of Gaussian noise enhances performance. We hypothesize that including a small amount of Gaussian noise smoothes the base probability density $p_0$ so that it remains well-defined over the higher-dimensional space. Note that this noise augmentation is only applied to $x_0$ but not to the conditioning information $z$, since the model relies on the precise conditioning information to construct the straight path.

\subsubsection{Latent Flow Matching}
In order to reduce the computational demands associated with training FM models for high-resolution image synthesis, we take inspiration from \cite{rombach2022high_latentdiffusion_ldm, dao2023flowlatent} and utilize an autoencoder model that provides a compressed latent space that aligns perceptually with the image pixel space similar to LDMs. By training in the latent space, we get a two-fold advantage: i) The computational cost associated with the training of flow-matching models is reduced substantially, thereby enhancing the overall training efficiency. ii) Leveraging the latent space unlocks the potential to synthesize images with significantly increased resolution efficiently and with a faster inference speed.

\subsection{High-Resolution Image Synthesis}
\label{sec:FM+DM}

Overall, our approach integrates all the components discussed above into a cohesive synthesis pipeline, as depicted in detail in \cref{fig:pipeline}. We start from a DM for content synthesis and move the generation to a latent space with a pretrained VAE encoder, which optimizes memory usage and enhances inference speed. To further alleviate the computational load of the DM and achieve additional acceleration, we adopt a relatively compact DM that produces compressed information. Subsequently, the FM model projects the compressed information to a high-resolution latent image with a straight conditional probability path. Finally, we decompress the latent space using a pre-trained VAE decoder. Note that the VAE decoder performs well across various resolutions, we show further proof in the appendix.

The integration of FM with DMs in the latent space presents a promising approach to address the trade-off between flexibility and efficiency in modeling the dynamic image synthesis process.
The inherent stochasticity within a DM's sampling process allows for a more nuanced representation of complex phenomena, while the FM model exhibits greater computational efficiency, which is useful when handling high-resolution images, but lower flexibility and image fidelity as of yet when it comes to image synthesis \cite{lipman2023flow}.
By combining them in the pipeline, we benefit from the flexibility of the DM while capitalizing on the efficiency of FM as well as a VAE.

%% file: figures/pipeline.tex
\begin{figure*}[th]
    \centering
    \begin{subfigure}[h]{\linewidth}
        \centering
        \includegraphics[scale=0.75]{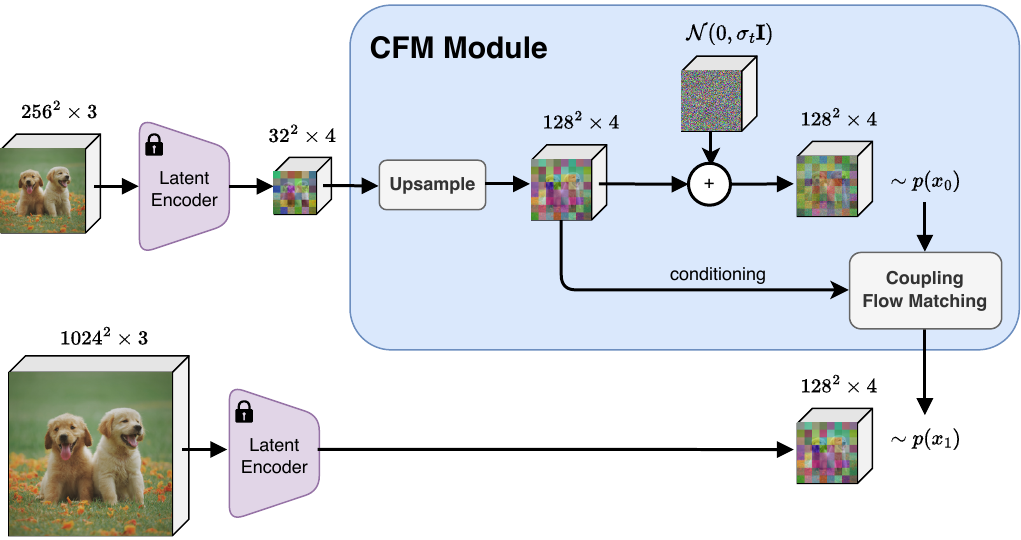}
        \vspace{-2mm}
        \caption{Training.}
        \label{fig:pipeline-training}
    \end{subfigure}
    \begin{subfigure}[h]{\linewidth}
        \centering
        \includegraphics[scale=0.75]{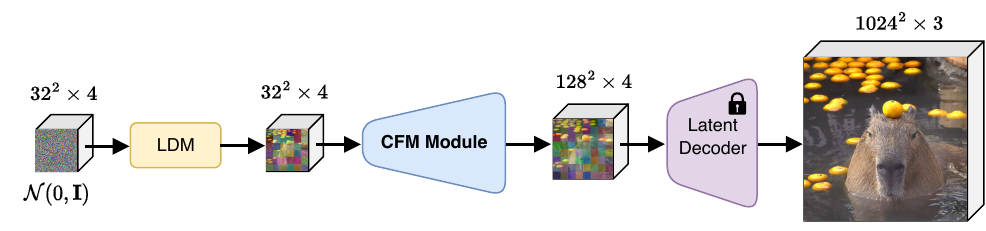}
        \vspace{-1mm}
        \caption{Inference.}
        \label{fig:pipeline-inference}
    \end{subfigure}
    \caption{Approach overview. \textbf{a)} During training we feed both a low- and a high-res image through the pre-trained encoder to obtain a low- and a high-res latent code, respectively. Based on the concatenated low-res latent code and a noisy version of it, the model regresses a vector field within $t \in [0, 1]$. \textbf{b)} During inference we can take any Latent Diffusion Model, generate the low-res latent, and then use our coupling flow matching model to synthesize the higher dimensional latent code. Finally, the pre-trained decoder projects the latent code back to pixel space.}
    \label{fig:pipeline}
    \vspace{-13pt}
\end{figure*}

%% file: figures/128_to_2k.tex
\begin{figure*}[h]
    \centering
    \scalebox{.95}{
    \includegraphics[width=\textwidth]{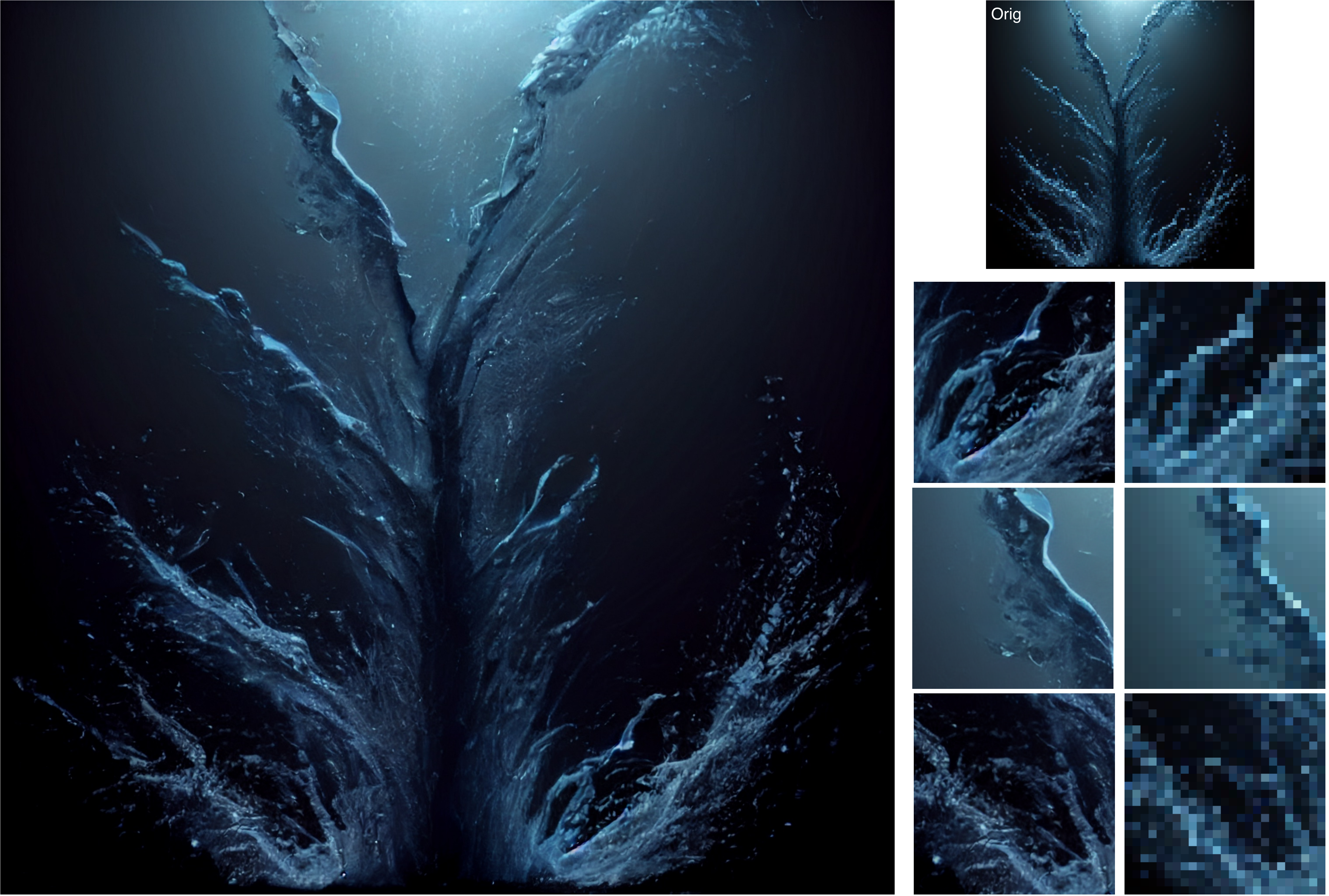}
    }
    \caption{Chaining our models enables elevating the image resolution from $128^2$ to $2048^2$ px. The contrast before and after upsampling is presented in the right column, with the original low-resolution image positioned in the top-right corner for reference. 
    }
    \label{fig:cover_4oct}
\end{figure*}

%% file: sec/4_Experiments.tex
\input{figures/uncurated_laion}

\subsection{Metrics and datasets}

For quantitative evaluation, we use the standard Fréchet Inception Distance (FID)\cite{heusel2017gans}, SSIM\cite{2004ssim}, and PSNR to measure the realism of the output distribution and the quality of the image. The general dataset we use for initial experiments and ablations is FacesHQ, a compilation of CelebA-HQ \cite{karras2017progressive_CelebAHQ} and FFHQ \cite{karras2019styleGAN}, as used in previous work~\cite{esser2021taming, saharia2022_SR3} for high-resolution synthesis tasks. However, as highlighted in \cite{chai2022anyresolution}, FID struggles to capture detail and measure fidelity at higher resolutions.
To remedy this, we also report p-FID \cite{heusel2017gans} for a more comprehensive evaluation, especially when images contain objects at different scales, such as LHQ~\cite{lhq}, which contains 90k high-resolution landscape images and offers a more diverse scale of scenes/objects presented in the image compared to FacesHQ. These two datasets serve as the basis for the evaluation.

For the general T2I image synthesis task, we train on the Unsplash dataset \cite{unsplash}, which provides diverse and high-quality images for training our model. Later, we evaluate on a high-resolution subset of LAION-5B~\cite{laion-5b} to check how well the model generalizes to unseen data.

\subsection{Boosting LDM with CFM}

\input{tables/laion_sampling}

Combining LDM with CFM achieves an optimal trade-off between computational efficiency and visual fidelity. We visualize the time taken by LDM and FM, respectively, to synthesize 1k resolution images in \cref{fig:barplot_nfe50_ldm-fm}, where LDM's inference time scales quadratically with increasing resolution, and inference is nearly impractical for real-time inference for a latent space of $128^2$. To ensure a fair comparison within the limited time frame, we compare our combination to the LCM-LoRA SDXL model \cite{song2023consistency, luo2023lcm}, which is known for its significantly faster inference than the original SDXL model. \cref{tab:laion_sampling} shows that our approach with a standard SD baseline model yields superior performance in terms of FID and inference speed. Note that we apply attention scaling \cite{jin_training-free_2023} on SD to synthesize images for varying resolutions and finetune the models, with more details in the Appendix.
We present a selection of image samples from the baseline SD1.5 model and CFM $64^2 \rightarrow 128^2$ in \cref{fig:uncurated-laion}.
We can equally upscale the LCM-LoRA SD1.5 model from 512 to 1k resolution images with our CFM model. We present our synthesized results in \cref{fig:LCM_SD1.5vsXL}. The inference time for a batch of four samples is $1.388$ seconds on an NVIDIA A100 GPU. The LCM-LoRA SDXL model fails to produce images with similar fidelity at the same resolution within the same time.

We further demonstrate the effectiveness of our approach by comparing it to state-of-the-art models \cite{zheng2024cogview3,pernias2023wuerstchen,podell2023sdxl} in image synthesis on COCO $1024\times1024$, including CogView3 \cite{zheng2024cogview3}. We reduce the computational cost of the diffusion component by using a lower resolution and fewer steps while offloading the remaining steps to our CFM module. This approach significantly reduces the inference time and maintains a good trade-off between speed and accuracy, as shown by the FID in \cref{tbl:result_coco}. In summary, we achieve a competitive FID at a faster inference speed than the counterpart diffusion models.

\input{tables/cogview3_tab}

\input{figures/timing_barplot_ldm-FM_nfe50}

\input{figures/2k}

\input{figures/faces_different_nfes}

\subsection{Baseline Comparison}

We compare our CFM model to three baseline methods on the FacesHQ and LHQ datasets. For a fair comparison, we fix the UNet architecture and hyperparameters so that the models only differ in their respective training objectives.

\input{figures/naive-upsampling}

\noindent \textbf{Regression Baseline}.
Similar to \cite{saharia2022_SR3}, we compare simple one-step regression models with $L1$ and $L2$ loss, respectively. The input is the low-resolution latent code and the target is the corresponding high-resolution latent code of the pre-trained KL autoencoder. In contrast, our method is trained with $L2$ loss on intermediate vector fields. \cref{tab:upsample_model} shows that CFM yields superior metric results. This is also reflected qualitatively, as visualized in \cref{fig:faces-model}, where the images from the regression baseline are visually blurry due to the mode-averaging behavior of the MSE regression. CFM excels at adding fine-grained, high-resolution detail to the image.
We conclude that simple regression models trained with $L1$ or $L2$ loss are not sufficient to increase resolution in latent space.

\input{tables/upsample_model}

\noindent \textbf{Diffusion Models}.
Based on optimal transport theory, the training of a constant velocity field presents a more straightforward training objective when contrasted with the intricate high-curvature probability paths found in DMs~\cite{lipman2023flow, dao2023flowlatent}. This distinction often translates to slower training convergence and potentially sub-optimal trajectories for DMs, which could detrimentally impact both training duration and overall model performance. \cref{fig:sr3-cfm-nfe-fid} shows that within $100$k iterations and for different numbers of function evaluations (NFE) after convergence, we consistently achieve a lower FID compared to the DM. In particular, the CFM model shows a faster reduction of the FID and provides better results.

\cref{tab:laion_cfm-vs-dm} shows that the combination of DM and CFM outperforms the cascaded DMs across the board.

Taken together, these results underscore the training efficiency of our CFM model over DMs and its superior performance on the up-sampling task after fewer training steps.

\input{figures/dm-cfm_FID-nfe-train}

\input{tables/laion_cfm-vs-dm}

\input{figures/noising_step_ablation}

\noindent \textbf{Na\"ive Flow Matching}.
Finally, we compare with Na\"ive Flow Matching (FM). Similar to DM, FM is conditioned on the low-resolution latent code and starts with Gaussian noise, but uses an optimal transport-based objective to regress the vector fields. In contrast, our CFM method starts directly from the low-resolution latent code and regresses the vector field to the high-resolution counterpart. Due to the presence of data-dependent coupling, we have guaranteed optimal transport during training. For all methods, the low-resolution latent code is available as conditioning information throughout the entire generation trajectory.
We evaluated the aforementioned two variants quantitatively (\cref{tab:upsample_model}) and qualitatively (\cref{fig:faces-model}), where we observed that the CFM model with data-dependent coupling readily outperforms the ones without.
We provide more information about the noise augmentation in \cref{fig:noising_steps_plot}. Notably, in the specific upsampling scenario from $256^2$ to $1024^2$, we observe an optimal configuration with a noising timestep of 400. The introduction of Gaussian noise proves beneficial as it imparts a smoothing effect on the input probability path, resulting in improved performance. However, excessive Gaussian noise can lead to the loss of valuable information, subsequently deteriorating the data-dependent coupling and reverting the model's behavior to FM's Gaussian assumption of $p(x_0)$. This finding underscores the delicate balance required in incorporating noise for optimal model performance.

\subsection{CFM for Degraded Image Super-Resolution}

Our model is originally intended to render image synthesis with existing diffusion models more effectively by enabling them to operate on a lower resolution while increasing pixel-level resolution. However, our method can also be generalized to work on super-resolution tasks which usually include image degradations \cite{wang2018esrgan} for low-resolution images. By fine-tuning our method, we can achieve state-of-the-art results on two common benchmark datasets on a $4 \times$ upsampling task from $128^2$ to $512^2$ pixels. We provide quantitative (\cref{tab:sota-sr_comparison}) and qualitative (\cref{fig:sup:div2k_sr}) results in the appendix.


\subsection{CFM Model Ablations}

\noindent \textbf{Upsampling Methods}
Since the dimensionality of the samples from both terminal distributions must be consistent for CFM, we need to upsample the low-resolution latent code $x_0$ to match the resolution at $x_1$. In this context, we perform an ablation study comparing three different upsampling methods: nearest neighbor upsampling, bilinear upsampling, and pixel space upsampling (PSU). The first two methods operate in latent space, while PSU requires the use of the KL autoencoder to upsample in pixel space. Denoting the latent encoder as $\mathcal{E}$, the decoder as $\mathcal{D}$, and the bilinear upsampling operation as $\textit{\text{UP}}$, the upsampling operation PSU can be represented as $\mathcal{E}(\textit{\text{UP}}(\mathcal{D}(\cdot)))$. We empirically find that upsampling in latent space works well, but introduces artifacts that make distribution matching with CFM more difficult. In contrast, PSU yields faster model convergence at minimal additional cost (see \cref{fig:sup:fid-time_upsampling}) and also makes our approach invariant to the autoencoder used. Therefore, we use PSU unless otherwise stated.

\noindent \textbf{Noise augmentation}
We systematically investigate the impact of varying levels of noise augmentation. \cref{fig:noising_steps_plot} shows the FID and Patch-FID for different noise augmentation steps, with higher values corresponding to more noise. Our findings suggest that noise augmentation is crucial for model performance, albeit being quite robust to the amount of noise. Empirically, we discovered that $t=400$ yields the best results overall.

\noindent\textbf{Intermediate Results along the ODE Trajectory} In \cref{fig:intermediate} we show intermediate results along the ODE trajectory. It can be seen that the CFM model gradually transforms the noise-augmented image representation to its high-resolution image counterpart.

%% file: figures/uncurated_laion.tex
\begin{figure*}
    \centering
    \includegraphics[width=\linewidth]{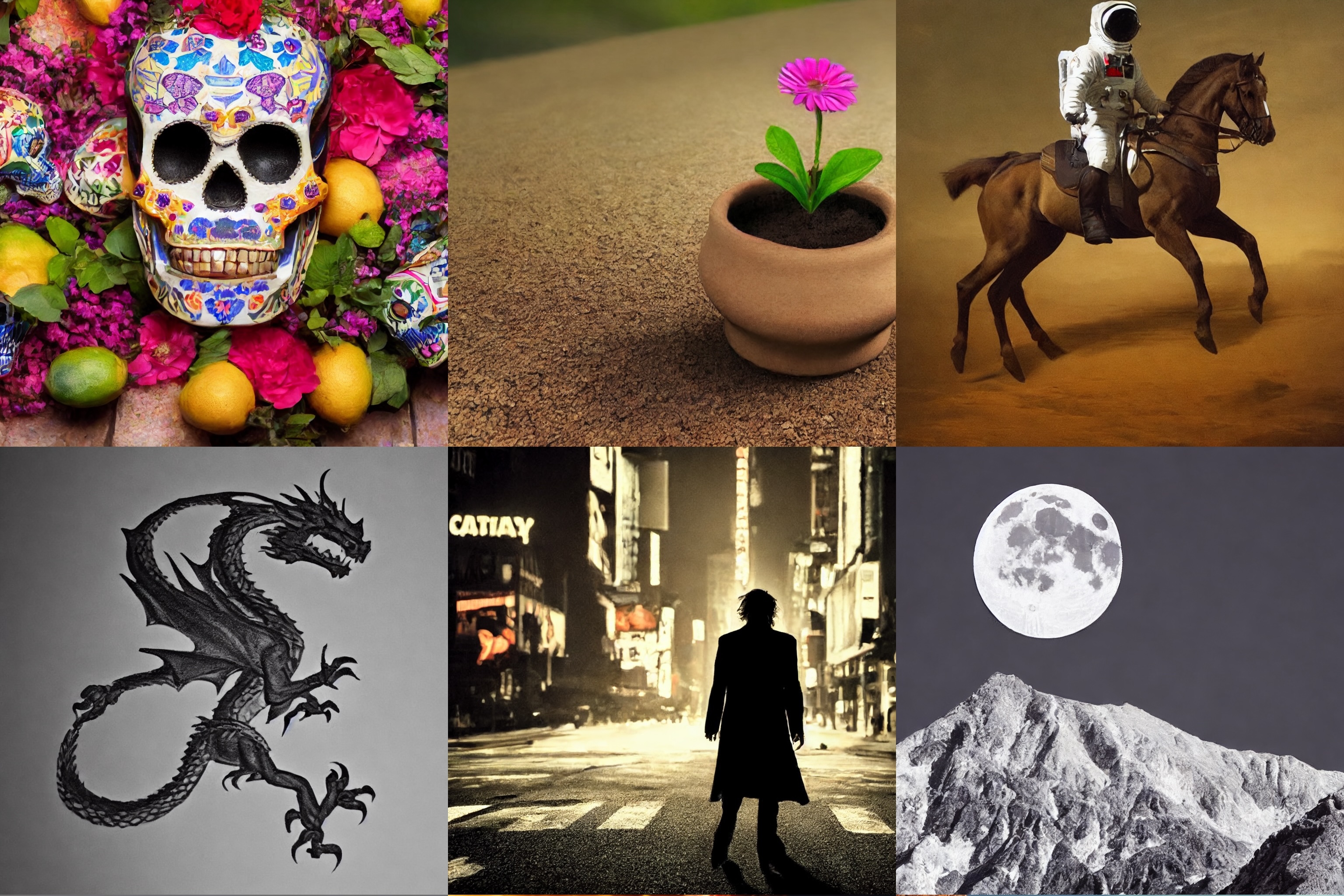}
    \caption{Uncurated samples from the Coupling Flow Matching model on top of SD 1.5 \cite{rombach2022high_latentdiffusion_ldm} using a classifier-free guidance scale of $7.5$. Samples are generated in \textit{latent space} $64^2$ and up-sampled with CFM from $64^2$ to $128^2$. The resulting images have a resolution of $1024 \times 1024$ pixels. Best viewed via zoomed in.}
    \label{fig:uncurated-laion}
\end{figure*}

%% file: tables/laion_sampling.tex
\begin{table*}
    \setlength\tabcolsep{10.pt}
    \centering

    \begin{tabular}{l||ccccc}
    \toprule
    \multicolumn{5}{l}{\bf{Zero-shot LAION-5k 1024$\times$1024}} \\
    \toprule
Model   & CLIP $\ua$   & FID $\da$  & p-FID $\da$  & time (\textit{s}/im) $\da$ \\
    \midrule
SD1.5 + \und{CFM} $\bf{256^2 \rightarrow 1024^2}$ 
        & 23.75        & \und{25.47}    & \und{23.31} & \bf{0.62} \\
SD1.5 + \und{CFM} $\bf{512^2 \rightarrow 1024^2}$ 
        & \bf{26.14}   & \bf{21.67}     & \bf{15.96}  & 3.16 \\
LCM-LoRA SDXL \cite{luo2023lcm}
        & \und{24.51}  & 28.98          & 24.00       & \und{1.83} \\
    \bottomrule
    \end{tabular}

\caption{Quantitative comparison for $1024^2$ image synthesis using SD v1.5 \cite{rombach2022high_latentdiffusion_ldm} plus our Coupling Flow Matching (CFM) method against a state-of-the-art diffusion speed-up method. The numbers after \und{CFM} symbolize the starting and ending resolutions in pixel space. FID and Patch-FID are computed for 5k samples. We use the fixed step-size Euler ODE solver with $40$ number of function evaluations for CFM. For LCM-LoRA SDXL \cite{luo2023lcm} we use $4$ sampling steps.}
\label{tab:laion_sampling}
\end{table*} 

%% file: tables/cogview3_tab.tex
\begin{table}[h]
    \begin{center}
    \begin{small}
    \setlength\tabcolsep{8pt}
    \begin{tabular}{lccc}
    \toprule
    \multicolumn{4}{l}{\bf{Zero-shot COCO-5k 1024$\times$1024 }} \\
    \toprule
    $\quad$ Model                       & Steps & Time Cost & FID$\downarrow$   \\
    \toprule
    $\quad$ SDXL~\cite{podell2023sdxl}  & 50    & 19.67s    & \bf{26.29}  \\
    $\quad$ StableCascade~\cite{pernias2023wuerstchen}               & 20+10 & 10.83s    & 36.59     \\
    $\quad$ CogView3~\cite{zheng2024cogview3}          & 50+10 & 10.33s    & 31.63      \\
    \midrule
    $\bf{256^2 \rightarrow 1024^2}$ & & &\\ 
    $\quad$ SD1.5 + \und{CFM}  & 25+20 & 4.07s & 33.48 \\
    $\quad$ SD1.5 + \und{CFM}  & 40+40 & 5.88s & 30.64 \\
    \midrule
    $\bf{512^2 \rightarrow 1024^2}$ & & &\\
    $\quad$ SD1.5 + \und{CFM}  & 25+20 & 8.79s & \und{28.81} \\
    \bottomrule
    \end{tabular}
    \caption{Metric results on 5k samples from the COCO dataset \cite{lin2014microsoft}. All samples are generated on $1024\times 1024$. The time cost is measured with a batch size of 4. Table adapted from \cite{zheng2024cogview3}.} %
    \label{tbl:result_coco}
    \end{small}
    \end{center}
\end{table}

%% file: figures/timing_barplot_ldm-FM_nfe50.tex
\begin{figure}
    \centering
    \includegraphics[width=0.4\textwidth]{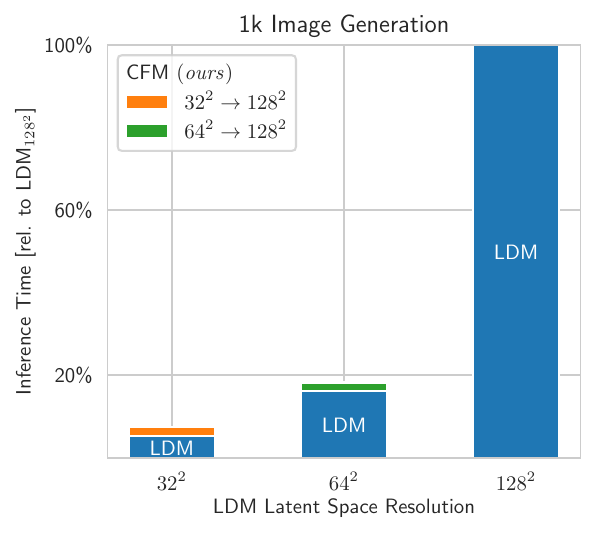}
    \caption{Comparison of 1k image synthesis performance using different architectures.
    We utilize SD v1.5 as our base model for LDM and adapt its resolution based on \cite{jin_training-free_2023}. LDM's inference time grows quadratically with higher resolutions, making real-time inference nearly impractical at a $128^2$ resolution latent space. In contrast, the integration of Coupling Flow Matching (CFM) with $50$ function evaluations exhibits consistently faster inference, highlighting its efficiency in high-resolution image synthesis.}
    \label{fig:barplot_nfe50_ldm-fm}
\end{figure}

%% file: figures/2k.tex
\begin{figure*}
    \centering
    \includegraphics[width=.84\textwidth]{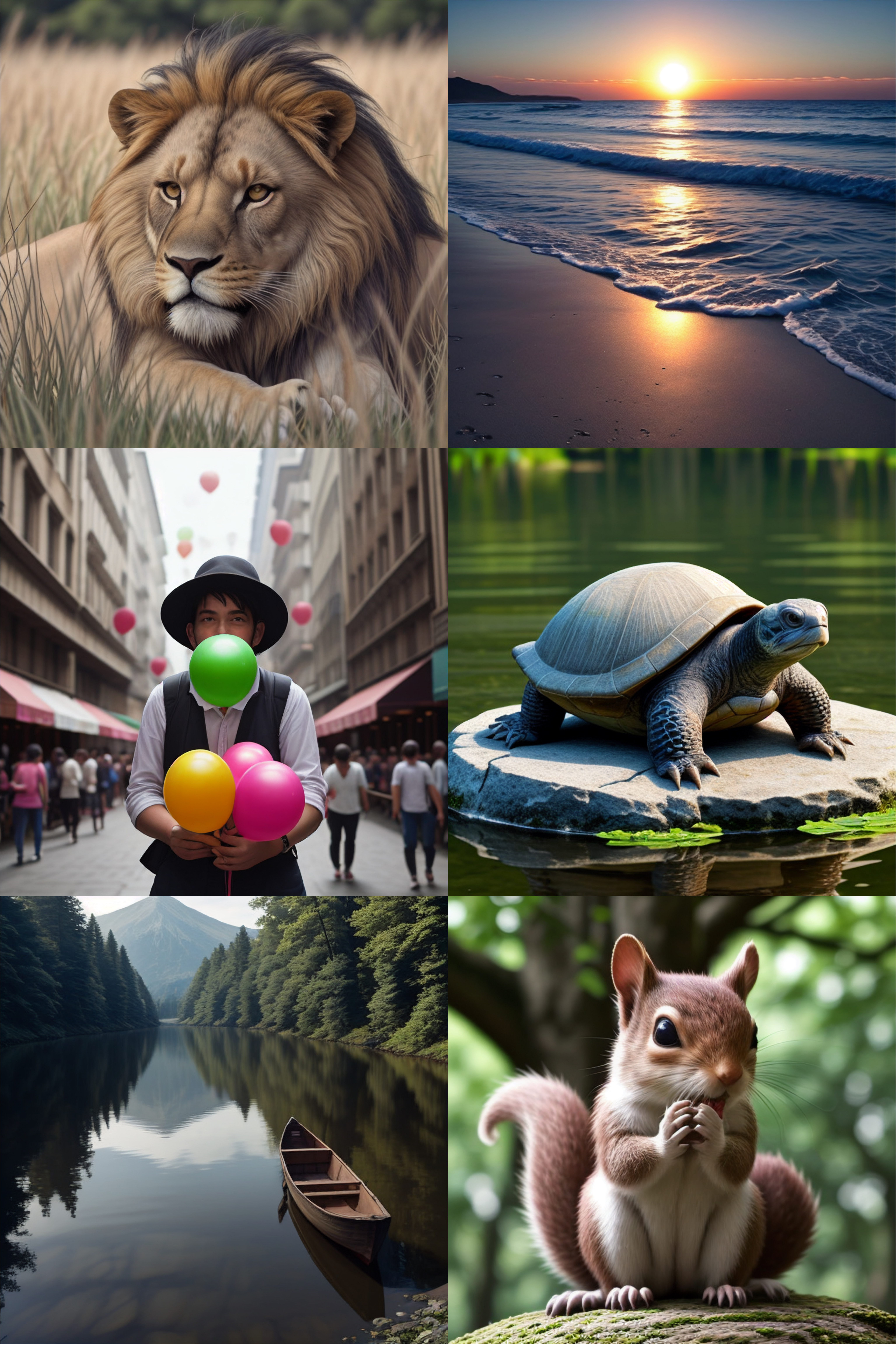}
    \caption{Samples synthesized in $2048^2$ px. The base diffusion model is SD1.5 synthesizing images in $512^2$ px. By plugging in our CFM module, we can quickly boost the resolution to 2k and generate high-fidelity images. Best viewed when zoomed in.
    }
    \label{fig:sup:2k_ensemble}
\end{figure*}

%% file: figures/faces_different_nfes.tex
\newlength{\mycwdddd}

\setlength{\mycwdddd}{ \dimexpr  0.166 \textwidth - 2 \tabcolsep  }
\begin{table*}[t]

\centering
\resizebox{.95\textwidth}{!}{
\begin{tabular}{ p{\mycwdddd} p{\mycwdddd}  p{\mycwdddd} p{\mycwdddd} p{\mycwdddd} p{\mycwdddd} }

    \centering GT & 
    \centering NFE=1 & 
    \centering NFE=2 & 
    \centering NFE=4 & 
    \centering NFE=10 & 
    \centering NFE=50 \tabularnewline

    \multicolumn{6}{p{0.98\textwidth  }}{\includegraphics[width=1.\linewidth]{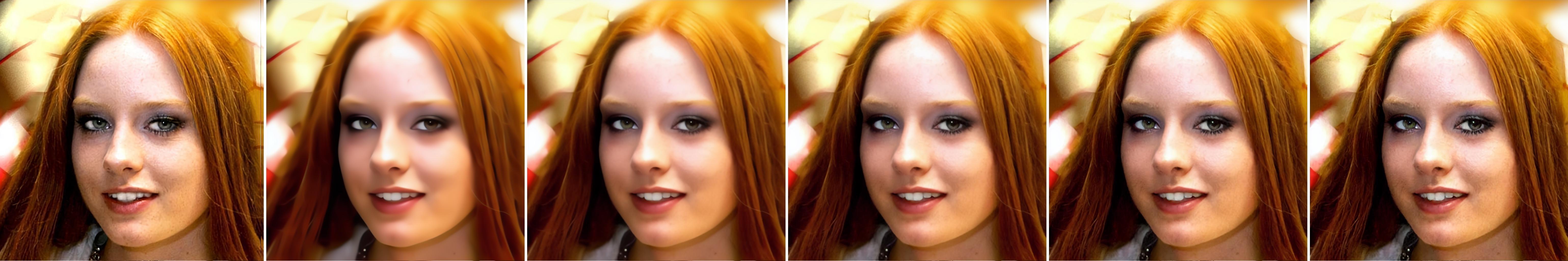}} 

\end{tabular}
}
\vspace{-2mm}
\captionof{figure}{
Sample quality for different number of function evaluations (NFE). From left to right, 1st column represents the ground truth, high-resolution image. From the 2nd column on, we show the results for NFE~$={1,2,4,10,50}$ with the Euler ODE solver.
}

\label{fig:num_NEF}
\end{table*}

%% file: figures/naive-upsampling.tex
\begin{figure*}[t]
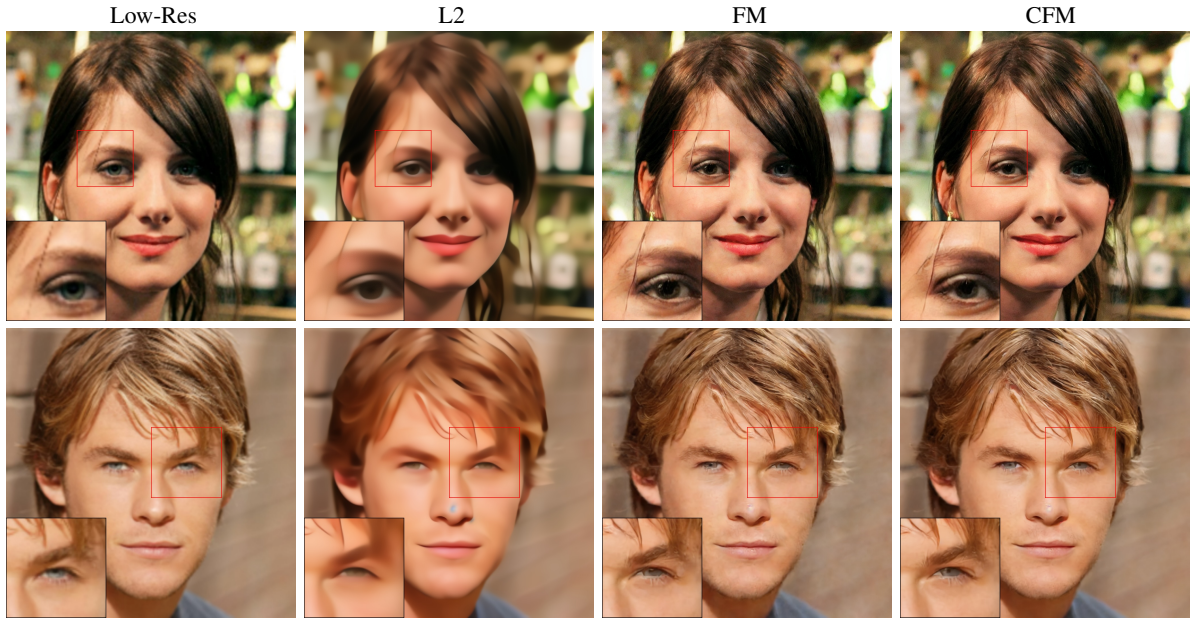

    \center \small
    \setlength\tabcolsep{0.6pt}
    \newcommand{\imgwidth}{0.22\textwidth}

    \newcommand{\imagepng}[2]{
    \includegraphics[width=\imgwidth]{figures/img/faces-baselines/#1_#2.jpg}
    }
    \newcommand{\rowy}[1]{
        \imagepng{#1}{lowres} & \imagepng{#1}{l2} & \imagepng{#1}{fm} & \imagepng{#1}{cfm}
    }
    \begin{tabular}{cccc}
        Low-Res  & L2 & FM & CFM \\
        \rowy{000057} \\
        \rowy{000024} \\
    \end{tabular}
    \vspace{-3mm}
    \caption{Results for different baseline methods, increasing resolution from $256^2$ px to $1024^2$ px. \textit{Low-Res} corresponds to bi-linear upsampling of the low-resolution image, \textit{L2} refers to the L2 regression baseline. \textit{FM} and \textit{CFM} correspond to Flow Matching and Coupling Flow Matching, respectively. Best viewed when zoomed in.
    }
    \label{fig:faces-model}
\end{figure*}

%% file: tables/upsample_model.tex
\begin{table*}[hbt!]
    \centering
    \begin{tabular}{l||cccc|cccc}
    \toprule
      & \multicolumn{4}{c|}{FacesHQ}                                           & \multicolumn{4}{c}{LHQ} \\
        Model               & SSIM$\uparrow$   & PSNR$\uparrow$   & FID$\downarrow$ & p-FID$\downarrow$   & SSIM$\uparrow$       & PSNR$\uparrow$       & FID$\downarrow$ & p-FID$\downarrow$  \\
        \midrule
        $L1$                & \textbf{0.86}     & \textbf{31.78}    & 4.52              & 6.51     & \textbf{0.72}         & \textbf{26.99}        & 4.88    & 6.54       \\
        $L2$                & \underline{0.85}  & \underline{31.48}                     & 5.73      & 9.07     & \textbf{0.72}         & \underline{26.87}                 & 6.02    & 8.59       \\
        DM                 & 0.73              & 23.68             & 2.72              & 4.71     & 0.61                  & 19.94                 & 4.29    & 4.55       \\
        FM                  & 0.82              & 30.46             & \underline{1.37}  & \underline{2.10}     & 0.68                  & 25.50                 & \underline{2.31}    & \underline{2.61}       \\
        \midrule
        CFM (\textit{ours}) & 0.82  & 30.40     & \textbf{1.36}     & \textbf{1.62}     & \underline{0.69}     & 25.69                 & \textbf{2.27}   & \textbf{2.38}   \\
    \bottomrule
    \end{tabular}
    \vspace{-2mm}
    \caption{Metric results for $L1$ and $L2$ regression, diffusion-based (\textit{DM}) similar to \cite{saharia2022_SR3}, Flow Matching (\textit{FM}) \cite{lipman2023flow}, and our Coupling Flow Matching (\textit{CFM}) on 5k samples from FacesHQ and LHQ high-resolution datasets, respectively.}
    \label{tab:upsample_model}
\end{table*}

%% file: figures/dm-cfm_FID-nfe-train.tex
\begin{figure*}
    \newcommand{\imgwidth}{0.75\linewidth}
    \newcommand{\subfigwidth}{0.49\linewidth}
    
    \centering
    \begin{subfigure}[h]{\subfigwidth}
        \centering
        \includegraphics[width=\imgwidth]{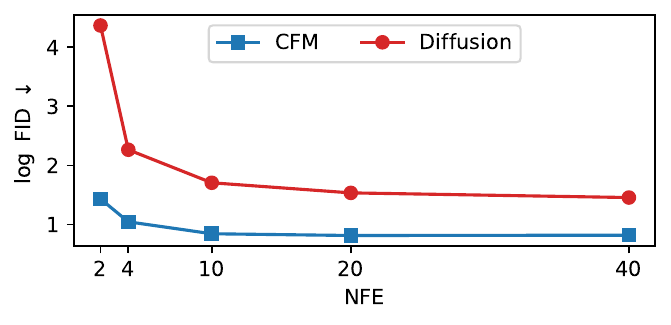}
        \includegraphics[width=\imgwidth]{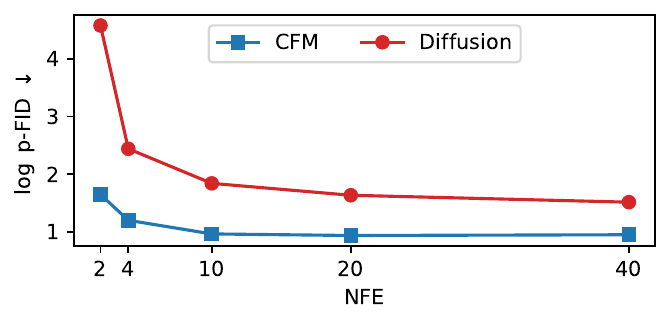}
        \label{}
    \end{subfigure}
    \begin{subfigure}[h]{\subfigwidth}
        \centering
        \includegraphics[width=\imgwidth]{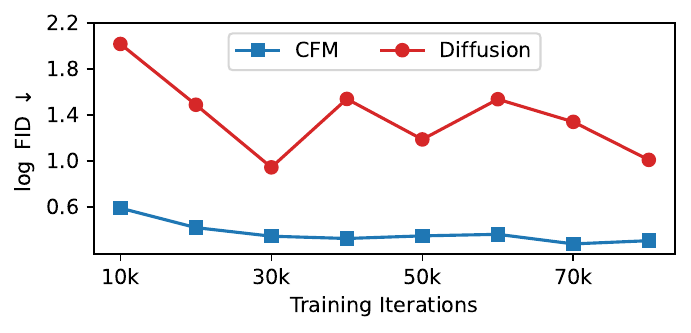}
        \includegraphics[width=\imgwidth]{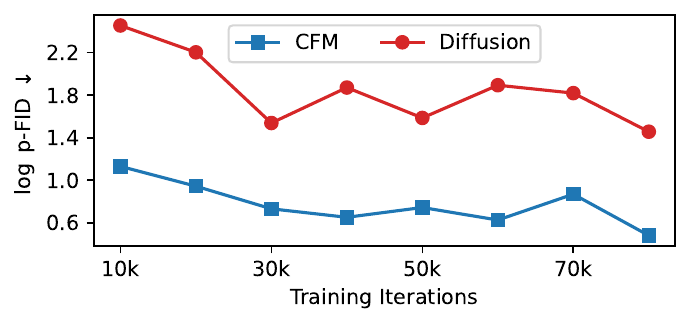}
        \label{}
    \end{subfigure}
    \vspace{-2.5mm}
    \caption{Comparison of a diffusion-based \cite{saharia2022_SR3} and our Coupling Flow Matching (CFM) module over the training for $4 \times$ up-sampling of the latent codes from $32^2 \rightarrow 128^2$. The decoded output resolution is $1024^2$. We report FID and p-FID for \textbf{a)} different numbers of function evaluations (NFE) and \textbf{b)} throughout training. Architecture and hyperparameters are kept fixed. FID evaluated on $5$k samples from the LHQ validation set. We use $50$ steps for both DDIM \cite{song2020denoising_ddim} sampling and the Euler ODE solver.
    }
    \label{fig:sr3-cfm-nfe-fid}
\end{figure*}

%% file: tables/laion_cfm-vs-dm.tex
\begin{table*}
    \setlength\tabcolsep{6pt}
    \centering
    \begin{tabular}{lc||cccc| cccc}
    \toprule
    \multicolumn{10}{l}{\bf{Zero-shot LAION-5k 1024$\times$1024}} \\
    \toprule
        & & \multicolumn{4}{c|}{$\bf{256^2 \rightarrow 1024^2}$ }         & \multicolumn{4}{c}{$\bf{512^2 \rightarrow 1024^2}$ } \\
        Model   & Steps & CLIP $\ua$    & FID $\da$ & p-FID $\da$   & time (\textit{s}/im) $\da$ & CLIP $\ua$    & FID $\da$ & p-FID $\da$   & time (\textit{s}/im) $\da$\\ 
    \midrule
    Diffusion
                & 4     & 22.92         &  35.22    & 50.53         & 0.19  & 25.68         &  26.53    & 25.20         & 2.72  \\
    CFM (\textit{ours})
                & 4     & 23.77         &  27.54    & 24.02         & 0.19  & 26.16         &  21.61    & 15.83         & 2.72   \\
    Diffusion
                & 40    & 23.55         &  26.67    & 24.16         & 0.62  & 26.05         &  22.29    & 16.36         & 3.16    \\
    CFM (\textit{ours})
                & 40    & 23.75         &  25.47    & 23.31         & 0.62  & 26.14         &  21.67    & 15.96         & 3.16   \\ \midrule
        \bottomrule
    \end{tabular}
    \vspace{-1.5mm}
 \caption{Quantitative comparison for $1024^2$ px image synthesis using SD1.5 \cite{rombach2022high_latentdiffusion_ldm} for sampling and either our Coupling Flow Matching (CFM) method or a diffusion-based latent space up-sampling model (DM) \cite{saharia2022_SR3}. FID and Patch-FID are computed for 5k samples. We use the Euler ODE solver for CFM and DDIM sampling for the DM.}
\label{tab:laion_cfm-vs-dm}
\end{table*}

%% file: figures/noising_step_ablation.tex

\begin{figure*}[hbt!]
    \centering
    \newcommand{\imgwidth}{0.75\linewidth}
    \centering
    \begin{subfigure}[h]{0.49\linewidth}
        \centering
        \includegraphics[width=\imgwidth]{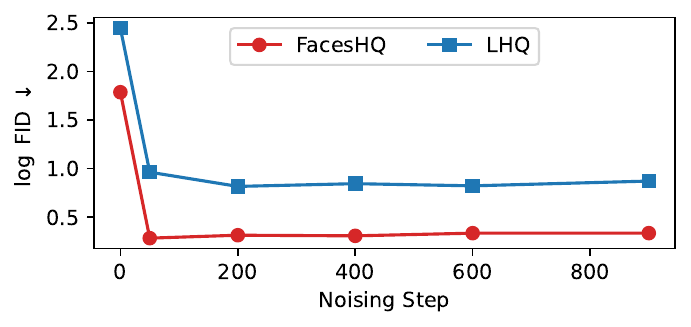}
    \end{subfigure}
    \begin{subfigure}[h]{0.49\linewidth}
        \centering
        \includegraphics[width=\imgwidth]{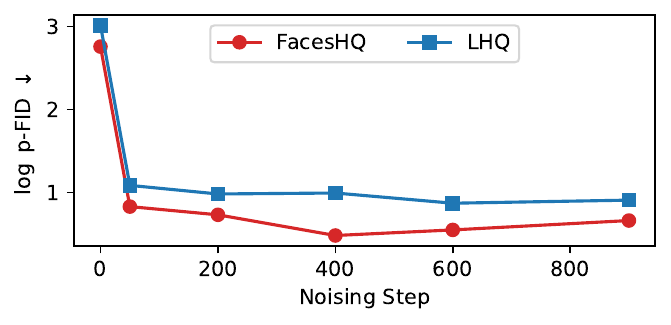}
    \end{subfigure}
    \vspace{-3mm}
    \caption{FID (\textit{left}) and Patch-FID (\textit{right}) for our model when applying different degrees of noise augmentation. Evaluated on $5$k samples.}
    \label{fig:noising_steps_plot}
\end{figure*}

%% file: sec/5_Conclusion.tex
Our work introduces a novel and effective approach to high-resolution image synthesis, combining the generation diversity of Diffusion Models, the efficiency of Flow Matching, and the effectiveness of convolutional decoders. Integrating Flow Matching models between a standard latent Diffusion model and the convolutional decoder enables a significant reduction in the computational cost of the generation process by letting the expensive Diffusion model operate at a lower resolution and up-scaling its outputs using an efficient Flow Matching model. 
Our Flow Matching model efficiently enhances the resolution of the latent space without compromising quality. 
Our approach complements DMs with their advancements and is orthogonal to their recent enhancements such as sampling acceleration and distillation techniques e.g., LCM \cite{luo2023lcm}. This allows for mutual benefits between different approaches and ensures the smooth integration of our method into existing frameworks.

%% file: sec/6_Appendix.tex
\renewcommand\thesection{\Alph{section}}
\setcounter{section}{0}

\section{Appendix for Boosting Latent Diffusion with Flow Matching}

\input{figures/sup/fid-time_faces_upsampling-method}

\subsection{Degraded Image Super-Resolution}

Originally, our model was designed to boost the resolution of existing diffusion models at a reduced cost. This differs from traditional super-resolution (SR) methods in two ways. First, they usually perform SR at lower resolutions, and second, they apply image degradation methods, \eg, compression artifacts, to obtain low-resolution images. Our model is not explicitly trained to be invariant to these degradations, but we can generalize to them. To support this claim, we additionally fine-tune our model on the Unsplash dataset, including image degradations following \cite{wang2021realesrgan} on a $4 \times$ upsampling task from $128^2$ to $512^2$ pixels. We then perform \textit{zero-shot} inference on the two common SR benchmark datasets DIV2K \cite{agustsson2017Div2k} and RealSR \cite{cai2019RealSR}.
\cref{tab:sota-sr_comparison} quantitatively shows the comparison to other state-of-the-art SR methods, where our method achieves comparable or even superior results in FID and CLIP-IQA. 
\cref{fig:nfe_stablesr} further compares the FID of our method against StableSR~\cite{wang2023stableSR} in the low number of function evaluations (NFEs) regime. Our model excels particularly at low NFEs due to its straighter trajectories compared to the diffusion-based counterpart.
\cref{fig:sup:div2k_sr} shows additional qualitative super-resolution results for mapping degraded low-resolution images to high-resolution images with our proposed CFM method.

\input{figures/cfm-stableSR_fid-nfe}

\input{tables/sota-sr_comparison}

\subsection{Additional Visualizations}

\subsubsection{Image Synthesis at 1024 and 2048 Resolutions} We present additional samples generated at resolutions of 1k and 2k from our pipeline in \cref{fig:sup:2k_ensemble,fig:sup:ldm64-generations,fig:2k-universe,fig:2k-eagle}. Our method produces high-resolution images with remarkable fidelity while maintaining fast inference times.

\subsubsection{Super-Resolution} \cref{fig:sup:div2k_sr} shows qualitative results of our method applied to the image super-resolution task. Low-resolution images include degradations following \cite{wang2021realesrgan} and we increase the resolution from $128^2$ to $512^2$ px. These results correspond to \cref{tab:sota-sr_comparison} from the main paper and showcase the generalizability of our model for super-resolution tasks.

\input{figures/sup/div2k_sr}

\input{figures/ODE_Trajectory}

\input{figures/sup/ldm64_generations}
\input{figures/sup/128_to_2k-universe}
\input{figures/sup/512_to_2k-eagle}

\subsection{Additional Ablations}

\subsubsection{Up-Sampling Method}

Our method requires matching resolutions between the source and target latent codes. We ablate three different options. The first two options involve directly upsampling the latent codes in the \emph{latent space}, either via bilinear or nearest neighbor upsampling. The third method pixel space upsampling (PSU) first decodes the image into \emph{pixel space}, then bilinearly upsamples the image, and encodes it back to latent space. This introduces some additional cost through the decoding and encoding operation, however, we found it to be negligible small.

\cref{tab:upsampling-comparison} shows that pixel space upsampling (PSU) performs particularly well. Besides performing better overall, PSU also mitigates the problem of artifacts for low-resolution images. \cref{fig:sup:psu-decoded} and \cref{fig:sup:autoencoder} show that the autoencoder introduces artifacts when encoding and decoding low-resolution images, e.g. of resolution $128^2$ and $256^2$ px. Increasing the resolution in pixel space and then encoding and decoding the image, avoids this issue. Since our primary goal is to speed up inference for image synthesis, using pixel space upsampling maintains most of the information present in the low-resolution latent code, which in turn ensures a well defined starting point for our Coupling Flow Matching module. This also makes the overall training more efficient as shown in

\input{figures/sup/psu_decoded}
\input{tables/upsampling-comparison}
\input{tables/autoencoder_metrics}

\subsubsection{Number of Function Evaluations}
\cref{tab:ode_step} shows the performance of our CFM model for different number of function evaluations. We evaluate the results on a subset of 1k samples of FacesHQ and present the total inference time on a single NVIDIA A100 GPU. We can clearly observe that our model achieves good performance already with as few as $10$ euler steps, indicating straight ODE trajectories.

\input{tables/supp_ode_step}

\subsubsection{Autoencoder Resolutions}
Even though the autoencoder from LDM \cite{rombach2022high_latentdiffusion_ldm} was trained on a fixed resolution, we find it to generalize well to images at different scales. This is shown quantitatively in \cref{tab:AE_metrics}, as well as qualitatively in \cref{fig:sup:autoencoder}.

\input{figures/sup/autoencoder_supp}

\subsection{Training Details}

The model architecture details for different datasets are provided in \cref{tab:architecture}, and we employ a learning rate of $5\times 10^{-5}$. We precompute the image latents to enhance computational efficiency. For fairness, we train the diffusion model counterpart for upsampling using the same architecture, utilizing 1000 diffusion steps and a cosine diffusion schedule.

\input{tables/fm_architecture}

\subsubsection{LDM $32^2$}

In the main paper we use a Latent Diffusion Model that operates on a lower-dimensional latent space of $32^2 \times 4$. The standard SD1.5 \cite{rombach2022high_latentdiffusion_ldm} was trained to work on latents with a dimensionality of $64^2$ pixels. After decoding the latent, this results in generated images with a resolution of $512^2$ pixels. Sampling latents of a different dimensionality is also possible but results in degraded performance, which is particularly pronounced for sampling lower-resolution latents. This deterioration can be partially mitigated by changing the scale of the self-attention layers from the standard $\sqrt{1/d}$ to $\sqrt{\log_TN/d}$ where $d$ is the inner attention dimensionality and $T$ and $N$ the number tokens during the training and inference phase respectively \cite{jin_training-free_2023}. 
We use this to rescale the attention layers of SD-1.5 for $32^2$ latents. Additionally, we finetune it on images of LAION-Aesthetics V2 6+ \cite{laion-aesthetics_v2} rescaled to $256^2$ pixels for one epoch, a learning rate of $1e{-5}$, and batch size of $256$. 

In \cref{tab:ldm256} we provide metrics for our fine-tuned Latent Diffusion Model on the LAION dataset. We find that FID plateaus with a classifier-free guidance scale at $7.0$. Consequently, we adopt this value for small resolution ($32^2$ in the latent space), text-guided image synthesis.

\input{tables/ldm256}

%% file: figures/sup/fid-time_faces_upsampling-method.tex
\begin{figure}[h]
    \centering
    \includegraphics[width=.95\linewidth]{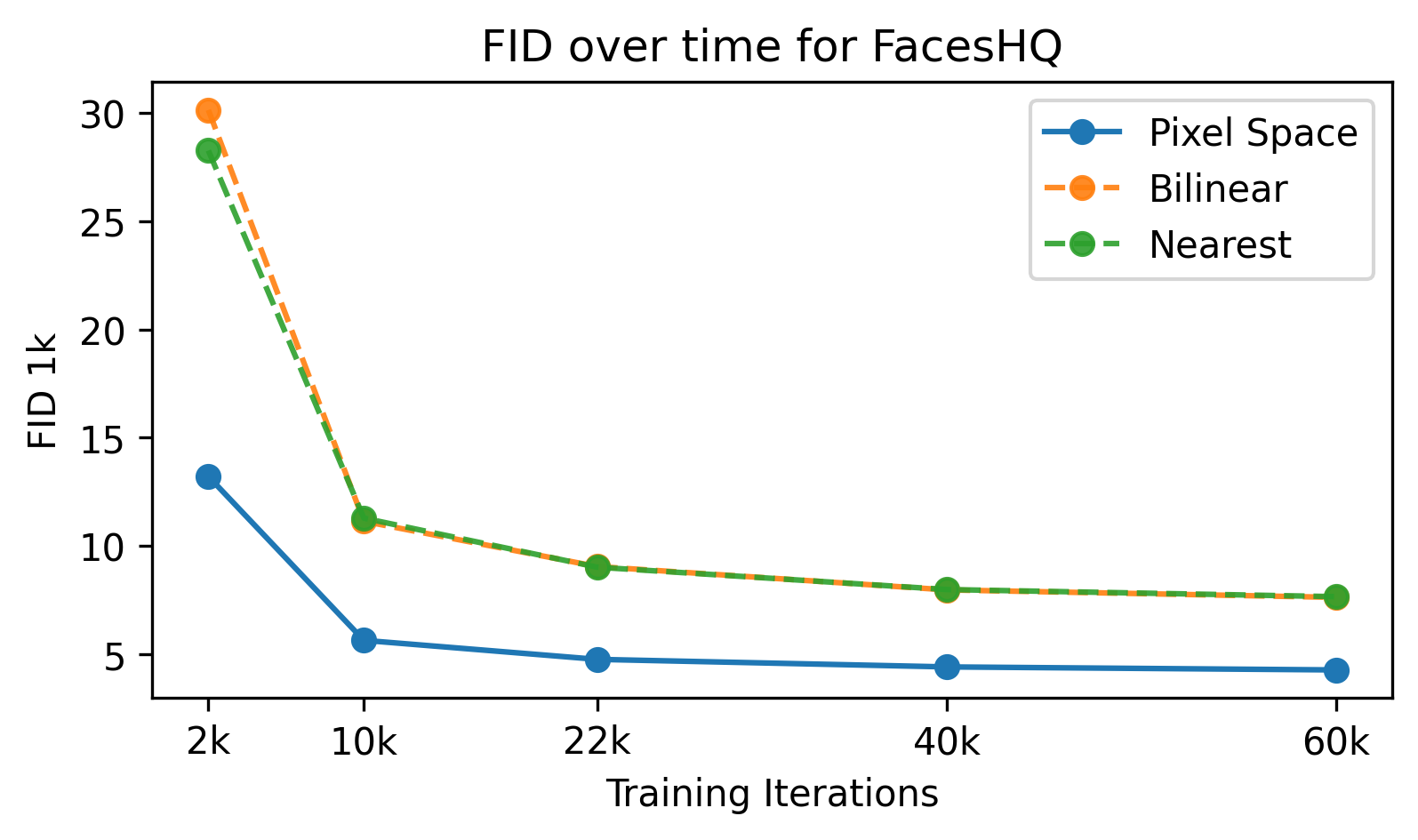}
    \caption{FID over training iterations for bilinear, nearest, and pixel-space upsampling of the low-resolution latent code on FacesHQ.}
    \label{fig:sup:fid-time_upsampling}
\end{figure}

%% file: figures/cfm-stableSR_fid-nfe.tex
\begin{figure}
    \newcommand{\imgwidth}{0.9\linewidth}
    \centering
    \includegraphics[width=\imgwidth]{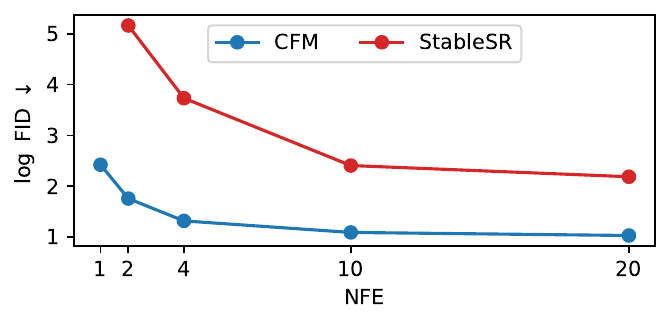}
    \vspace{-3mm}
    \caption{In the low NFE regime we outperform StableSR~\cite{wang2023stableSR} on FID at the $128^2\rightarrow 512^2$ px super-resolution task.}
  \label{fig:nfe_stablesr}
  \vspace{-4mm}
\end{figure}

%% file: tables/sota-sr_comparison.tex
\begin{table*}[t]
    \centering
    \begin{tabular}{l||cccc|cccc}
    \toprule
            & \multicolumn{4}{c|}{DIV2K~\cite{agustsson2017Div2k}}                                & \multicolumn{4}{c}{RealSR~\cite{cai2019RealSR}} \\
Model   & SSIM$\ua$   & PSNR$\ua$   & FID$\da$     & CLIP-IQA$\ua$   & SSIM$\ua$    & PSNR$\ua$    & FID$\da$     & CLIP-IQA$\ua$ \\
    \midrule
BSRGAN
        & 0.63         & \und{24.60}  & 44.22         & 0.52            & \und{0.77}    & 23.69         & 141.28        & 0.50      \\
RealERSGAN
        & \bf{0.64}    & 24.33        & 37.64         & 0.52            & 0.76          & 25.69         & 135.18        & 0.44      \\
DASR
        & \und{0.63}   & 24.50  & 49.16         & 0.50            & \bf{0.77}     & \bf{27.02}    & 132.63        & 0.31      \\
LDM
        & 0.58         & 23.36        & 36.21         & 0.62            & 0.72          & 25.49         & 132.32        & 0.59      \\
StableSR
        & 0.57         & 23.31        & \bf{24.67}    & \und{0.67}      & 0.70          & 24.69         & \und{127.20}  & \und{0.62}    \\
ResShift
        & 0.62         & \bf{24.69}   & 36.01         & 0.61            & 0.74          & \und{26.31}   & 142.81        & 0.55      \\\midrule
CFM (\textit{ours})
        & 0.52         & 21.63        & \und{29.76}   & \bf{0.73}       & 0.65          & 23.31         & \bf{125.88}   & \bf{0.67}    \\
    \bottomrule
    \end{tabular}
\caption{Quantitative comparison to other State-of-the-Art Super-Resolution models on two benchmark datasets. Comparison models are BSRGAN~\cite{zhang2021bsrgan}, RealERSGAN~\cite{wang2021realERSGAN}, DASR~\cite{liang2022DASR}, LDM~\cite{rombach2022high_latentdiffusion_ldm}, StableSR~\cite{wang2023stableSR}, and ResShift~\cite{yue2024resshift}.}
\label{tab:sota-sr_comparison}
\end{table*}

%% file: figures/sup/div2k_sr.tex
\begin{figure*}[t]
\setlength\tabcolsep{0.5pt}
\center
\small

\newcommand{\imgwidth}{0.2\textwidth}

\newcommand{\image}[2]{
    \includegraphics[width=\imgwidth]{figures/sup/div2k_sr/#1_#2.png}
}

\newcommand{\rowSR}[1]{\image{#1}{hr} & \image{#1}{lr} & \image{#1}{pred}}

\begin{tabular}{ccc}
    Ground Truth & Degraded Low-Res & CFM Prediction \\
    \rowSR{000019} \\       
    \rowSR{000020} \\       
    \rowSR{000043} \\
    \rowSR{000044} \\
    \rowSR{000077} \\
    \rowSR{000109} \\       

\end{tabular}
\caption{Qualitative results for image super-resolution on degraded images from $128^2$ to $512^2$ px with our CFM model.}
\label{fig:sup:div2k_sr}
\end{figure*}

%% file: figures/ODE_Trajectory.tex
\begin{figure*}    
    \centering
    \includegraphics[width=0.95\linewidth]{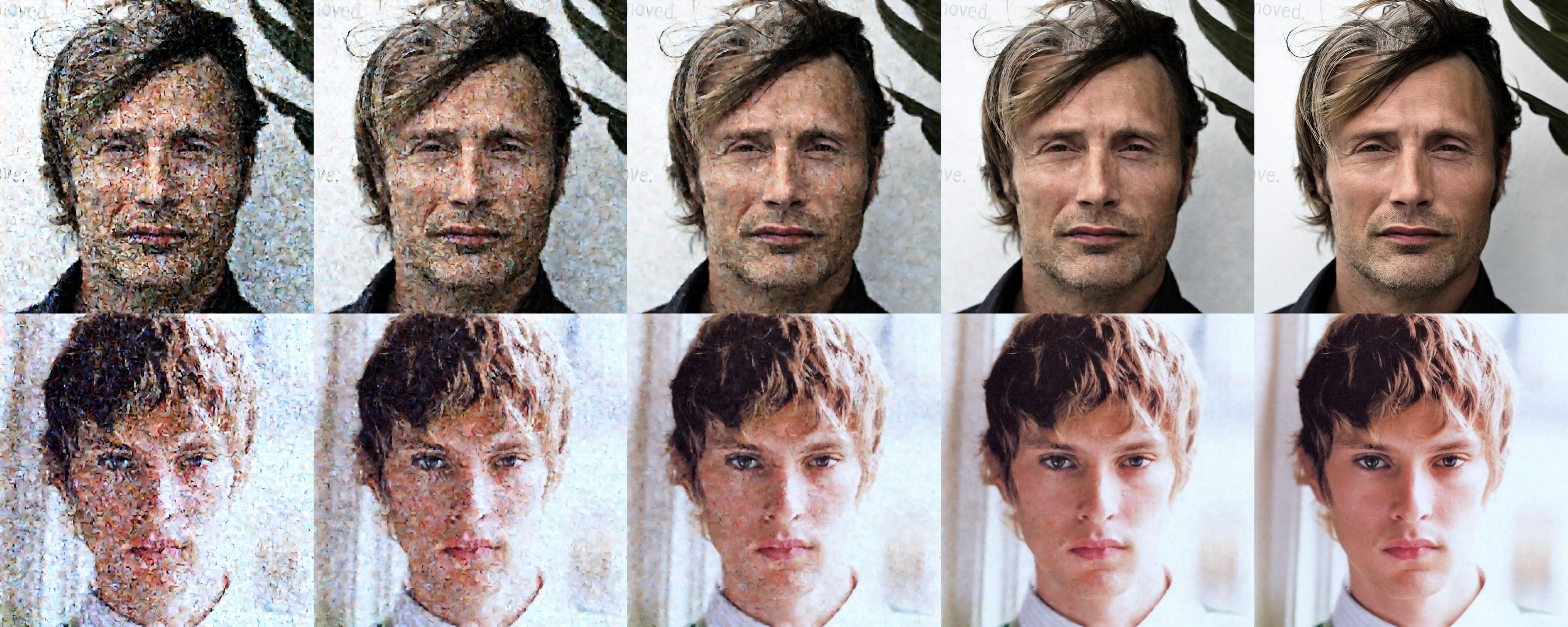}
    \includegraphics[width=0.95\linewidth]{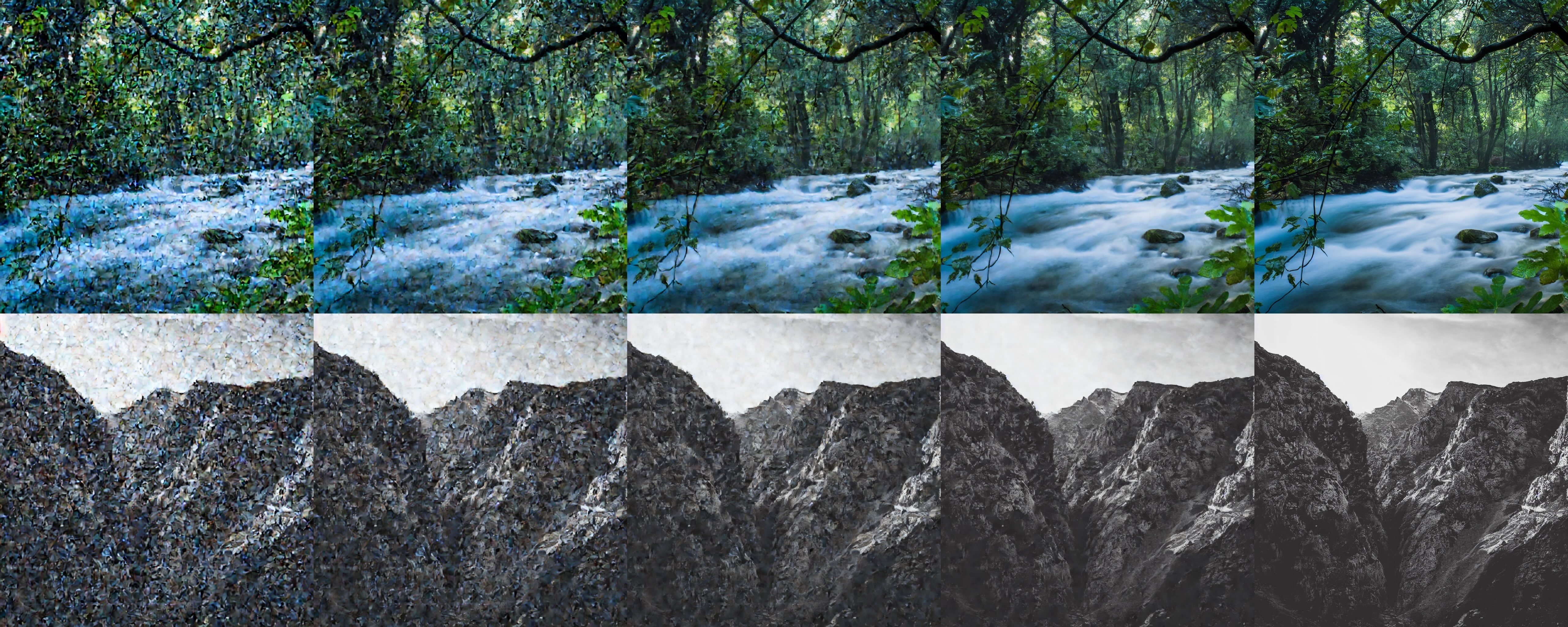}
    \includegraphics[width=0.95\linewidth]{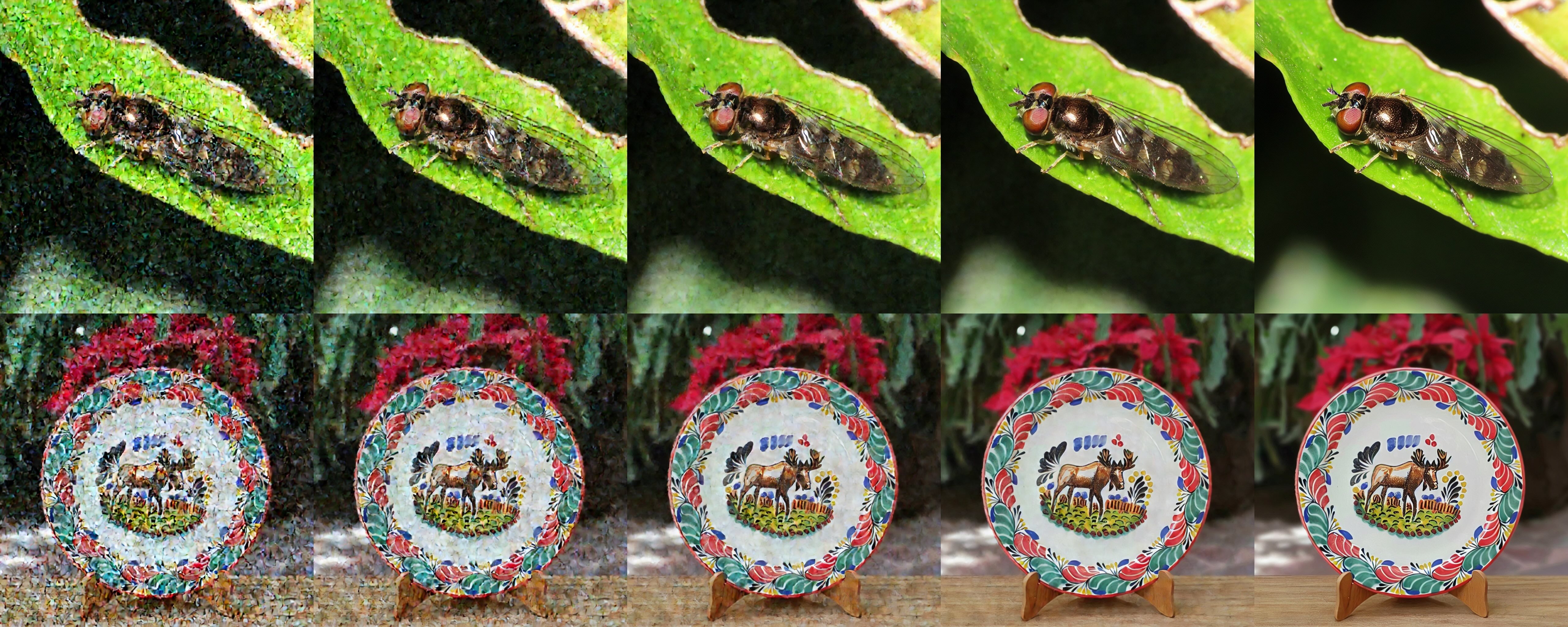}
    \caption{Samples decoded along the ODE trajectory for FacesHQ, LHQ, and LAION at $t \in \{0, 0.25, 0.5, 0.75, 1\}$, with a total number of function evaluations of 100. Best viewed when zoomed in.}
    \label{fig:intermediate}
\end{figure*}

%% file: figures/sup/ldm64_generations.tex
\begin{figure*}
    \centering
    \includegraphics[width=.83\textwidth]{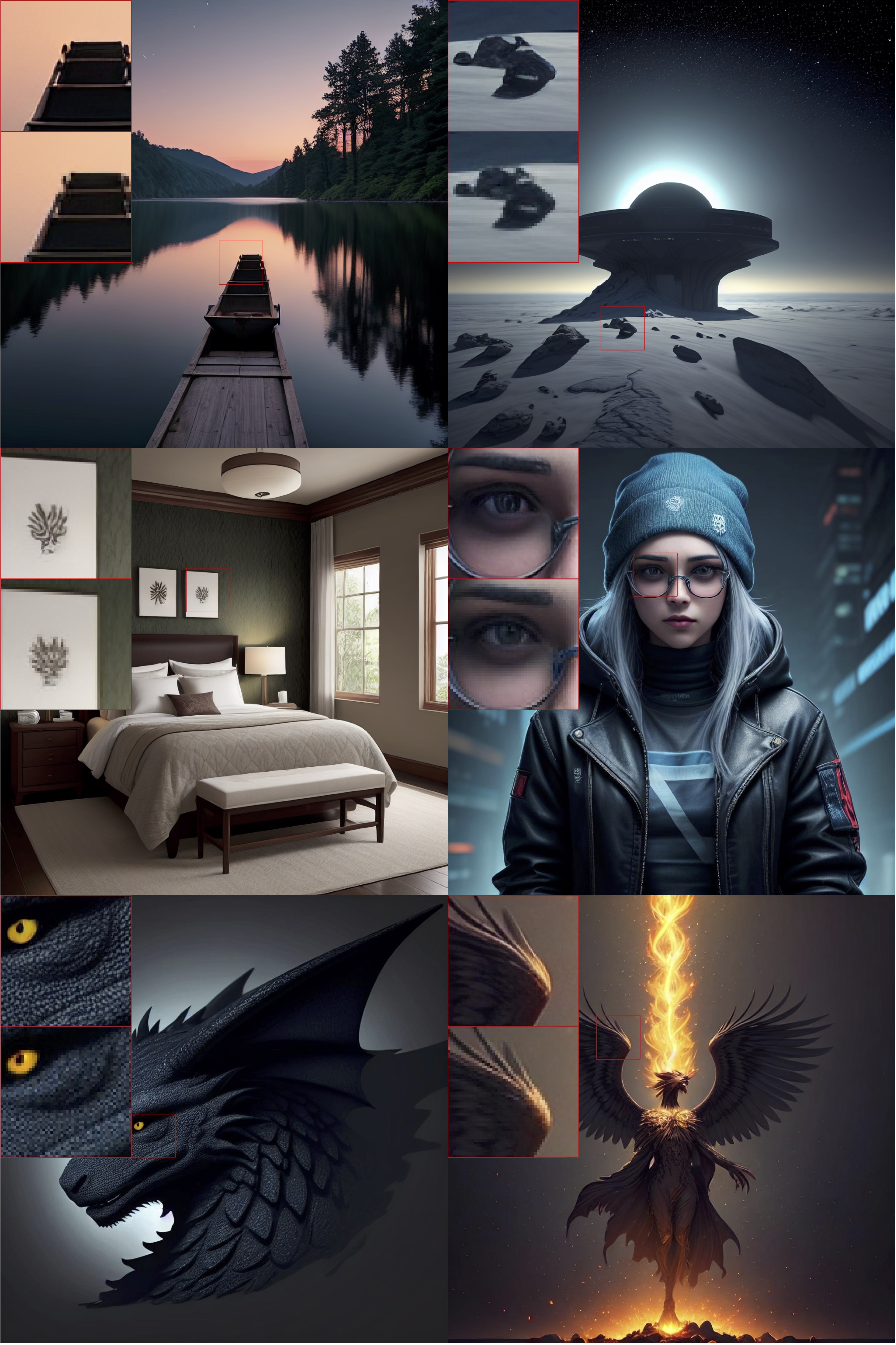}
    \caption{Samples synthesized in $1024^2$ px. The comparison is highlighted on the top-left corners between the samples generated solely from DM and those generated from the combination between DM and CFM.}
    \label{fig:sup:ldm64-generations}
\end{figure*}

%% file: figures/sup/128_to_2k-universe.tex
\begin{figure*}[h]
    \centering
    \includegraphics[width=.8\textwidth]{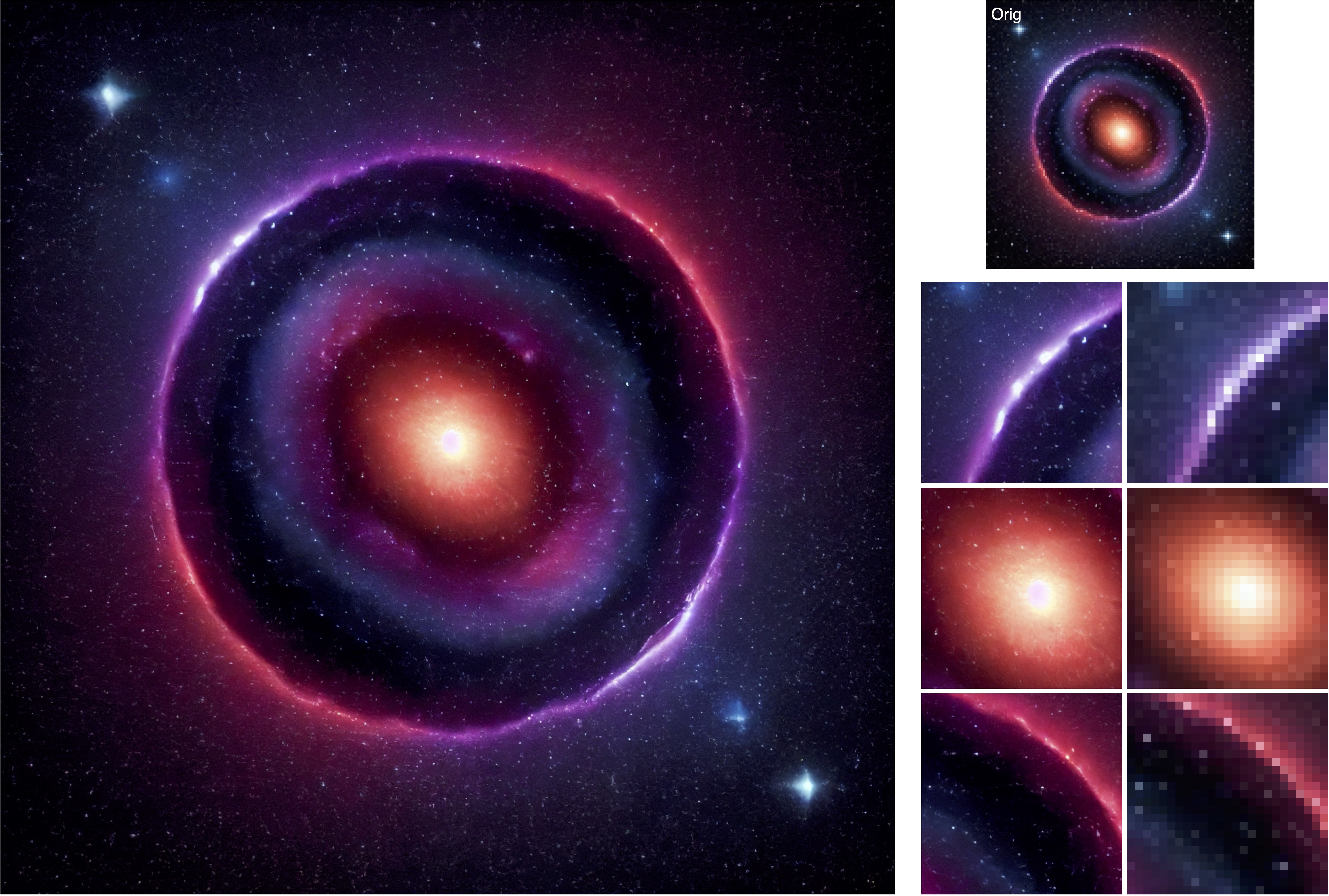}
    \caption{Chaining our models enables elevating the image resolution from $128^2$ to $2048^2$ px. The contrast before and after upsampling is presented in the right column, with the original low-resolution image positioned in the top-right corner for reference. 
    }
    \label{fig:2k-universe}
\end{figure*}

%% file: figures/sup/512_to_2k-eagle.tex
\begin{figure*}[h]
    \centering
    \includegraphics[width=.8\textwidth]{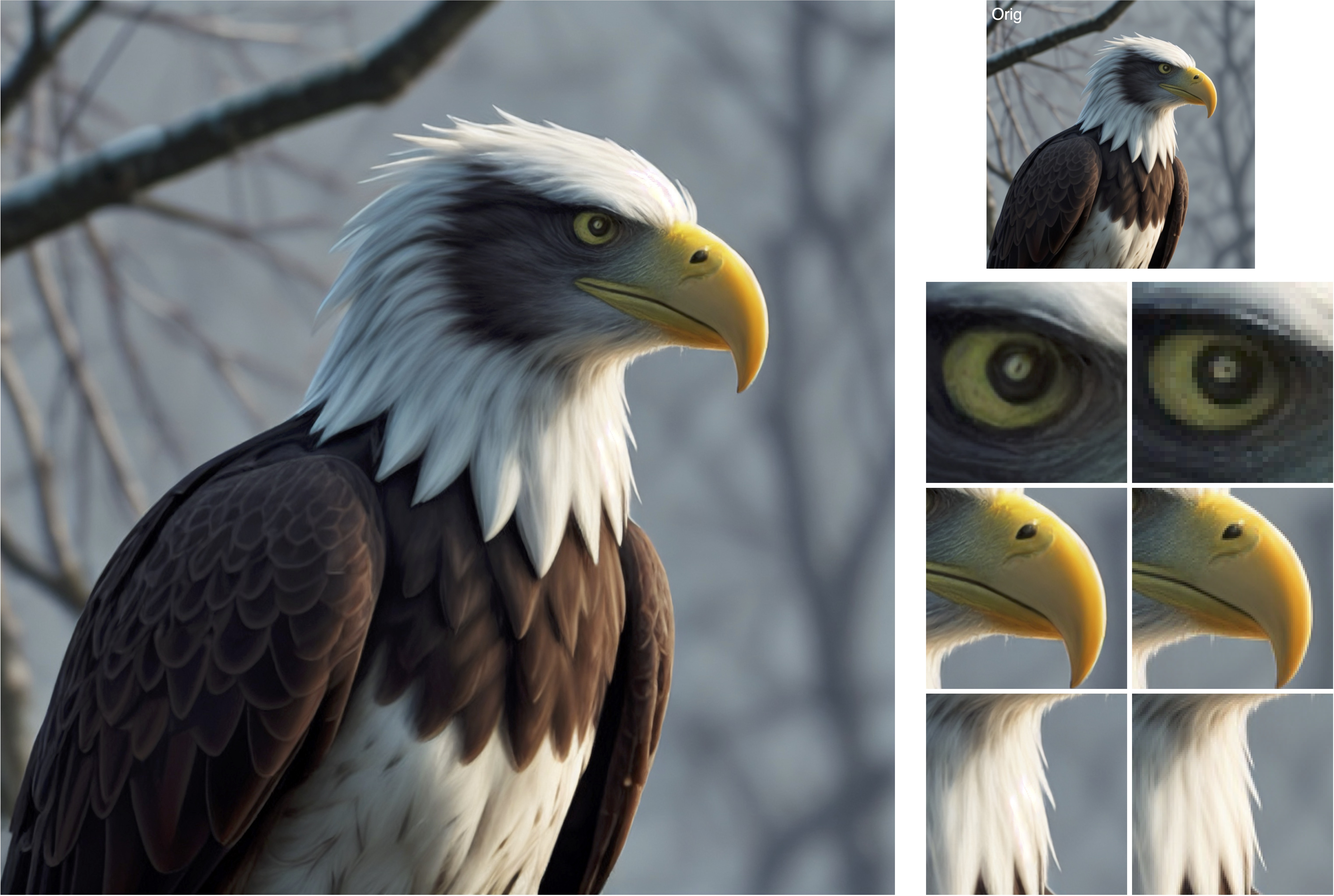}
    \caption{Upsampling results from $512^2$ to $2048^2$ px. The contrast before and after upsampling is presented in the right column, with the original low-resolution image positioned in the top-right corner for reference.}
    \label{fig:2k-eagle}
\end{figure*}

%% file: figures/sup/psu_decoded.tex
\begin{figure*}[h]
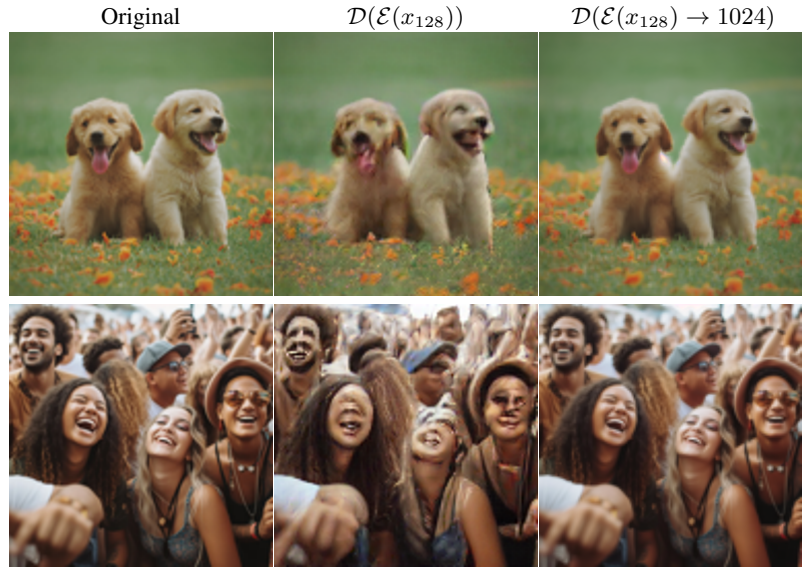

    \setlength\tabcolsep{0.4pt}
    \center
    \small
    \newcommand{\imgwidth}{0.2\textwidth}
    
    \newcommand{\imagepng}[2]{\includegraphics[width=\imgwidth]{figures/sup/psu-nearest-bilinear/#1_#2.png}}
        
    \newcommand{\rowimages}[1]{
        \imagepng{#1}{orig} & \imagepng{#1}{dec} & \imagepng{#1}{dec_psu}
    }
    
    \begin{tabular}{ccc}
        Original & $\mathcal{D}(\mathcal{E}(x_{128}))$ & $\mathcal{D}(\mathcal{E}(x_{128}) \rightarrow 1024)$ \\
        \rowimages{dogs} \\
        \rowimages{faces} \\
    \end{tabular}
    \caption{\textit{Left}: Original image with $128^2$ px resolution, bilinearly upsampled to $1024^2$ px. \textit{Middle}: $128^2$ px image encoded and decoded with the autoencoder, clearly showing artifacts. \textit{Right}: $128^2$ px image bilinearly upsampled to $1024^2$ px in pixel space and then encoded and decoded with the autoencoder. Besides being blurry, the decoded image shows no artifacts.}
    \label{fig:sup:psu-decoded}
\end{figure*}

%% file: tables/upsampling-comparison.tex
\begin{table*}[]
\centering
\begin{tabular}{@{}l||ccccc|ccccc@{}}
\toprule
\multicolumn{1}{l||}{} & \multicolumn{5}{c|}{FacesHQ}                                                                   & \multicolumn{5}{c}{LHQ}                                                                        \\
Upsampling            & PSNR$\uparrow$   & SSIM$\uparrow$  & MSE$\downarrow$ & FID$\downarrow$ & p-FID$\downarrow$ & PSNR$\uparrow$   & SSIM$\uparrow$  & MSE$\downarrow$ & FID$\downarrow$ & p-FID$\downarrow$ \\ \midrule
Bilinear              & 25.68          & 0.71          & 0.012          & 3.30          & 3.96                & 23.07          & 0.60          & 0.036         & 3.81          & 4.13                \\
Nearest               & 25.55          & 0.71          & 0.013          & 3.32          & 3.77                & 22.81          & 0.59          & 0.038          & 4.03          & 4.67                \\
Pixel Space      & \textbf{30.40} & \textbf{0.82} & \textbf{0.004} & \textbf{1.35} & \textbf{1.61}       & \textbf{25.49} & \textbf{0.68} & \textbf{0.022} & \textbf{2.32} & \textbf{2.70}       \\ \bottomrule
\end{tabular}
\caption{Ablation of upsampling methods for our Coupling Flow Matching model in the latent space. The \textit{pixel space upsampling} (PSU) method exhibits constantly better results than the latent space upsampling methods.}
\label{tab:upsampling-comparison}
\end{table*}

%% file: tables/autoencoder_metrics.tex
\begin{table}[htbp]
    \vspace{-4mm}
    \centering   
    \setlength\tabcolsep{4pt}
    \begin{tabular}{c||ccc}
    \toprule
      & \multicolumn{3}{c}{LAION-10k} \\
    \textbf{Image Size} & \textbf{SSIM} $\uparrow$ & \textbf{PSNR} $\uparrow$ & \textbf{FID} $\downarrow$ \\
    \midrule
    256    & 0.82   &  26.42   &   2.47  \\
    512     & 0.86  & 28.65    &  1.28  \\
    1024   & 0.88   &  30.72  & 0.84 \\
    2048   & 0.88 &  32.20  &  0.60 \\
    \bottomrule
    \end{tabular}
    \caption{Evaluation of the pre-trained autoencoder for different image resolutions.}
    \label{tab:AE_metrics}
\end{table}

%% file: tables/supp_ode_step.tex
\begin{table}
    \centering   
    \begin{tabular}{c||cccc}
    \toprule
      & \multicolumn{4}{c}{FacesHQ-1k} \\
    \textbf{NFE} & \textbf{SSIM} $\uparrow$ & \textbf{PSNR} $\uparrow$ & \textbf{FID} $\downarrow$ & \textbf{Time} (ms/img) $\downarrow$ \\
    \midrule
    1    & 0.81   &  27.41   &   66.24  & {\color{white}0,}395\\
    2     & 0.81  & 27.86    &  33.89  & {\color{white}0,}399 \\
    4   & 0.79   &  27.71  & 17.69  & {\color{white}0,}413 \\
    10   & 0.78 &  27.10  &  11.53  & {\color{white}0,}465\\
    50   & 0.75 &  26.50  &  10.31  & {\color{white}0,}718\\
    100   & 0.75 &  26.50 &  10.30  & 1,046\\
    \midrule
    dopri5 & 0.75 & 26.31 & 10.34 & {\color{white}0,}996 \\
    \bottomrule
    \end{tabular}
    \caption{We ablate the CFM-400 model and compare different numbers of function evaluations (NFE) during inference for the \textit{Euler} method, as well as the adaptive step-size Dormand-Prince solver.} 
    \label{tab:ode_step}
\end{table}

%% file: figures/sup/autoencoder_supp.tex
\begin{figure*}[h]
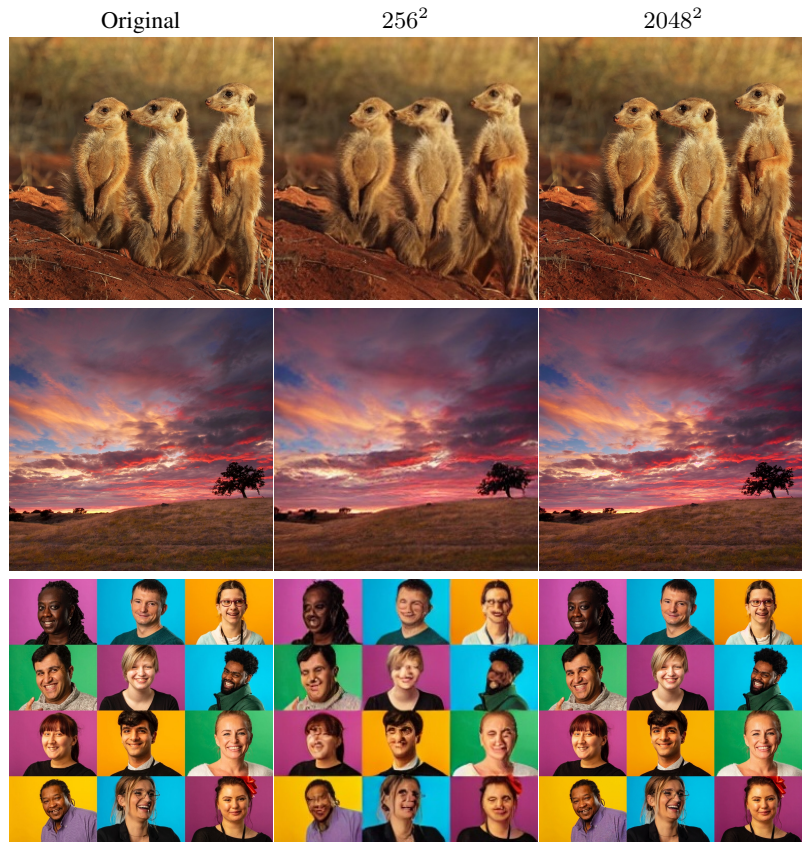


    \setlength\tabcolsep{0.4pt}
    \center
    \small
    \newcommand{\imgwidth}{0.2\textwidth}
    
    \newcommand{\imagepng}[2]{\includegraphics[width=\imgwidth]{figures/sup/autoencoder/#1#2_gen.jpg}}
        
    \newcommand{\rowimages}[1]{
        \imagepng{#1}{1024} & \imagepng{#1}{256} & \imagepng{#1}{2048}
    }
    
    \begin{tabular}{ccc}
        Original & $256^2$ & $2048^2$ \\
        \rowimages{ani} \\
        \rowimages{dawn} \\
        \rowimages{face} \\
    \end{tabular}
    \caption{Comparison between reconstructed images and their high-resolution original input using the same pre-trained autoencoder.  We can observe that the (i) pre-trained autoencoder can encode and decode images at different scales. (ii) It cannot reconstruct faces correctly in low resolution. (iii) The artifacts diminish with a higher resolution. Best viewed when zoomed in.}
    \label{fig:sup:autoencoder}
\end{figure*}

%% file: tables/fm_architecture.tex
\begin{table}
\centering 
    \vspace{-4mm}
    \begin{tabular}{lccc}
    \toprule
    \multicolumn{1}{c}{}  & \textbf{FacesHQ}    &\textbf{LHQ}       & \textbf{Unsplash}\\ \midrule

    Model size            & 113M                & 306M              & 306M          \\
    Channels              & 128                 & 128               & 128           \\
    Depth                 & 3                   & 3                 & 3             \\
    Channel multiplier    & 1, 2, 3, 4          & 1, 2, 4, 8        & 1, 2, 4, 8    \\
    Attention resolutions & 16                  & 16                & 16            \\
    Head channels         & 64                  & 64                & 64            \\
    Number of heads       & 4                   & 4                 & 4             \\ 
    Batch size            & 96                  & 128               & 768           \\

    \bottomrule
    \end{tabular}
\caption{Hyperparameters and number of parameters for our Coupling Flow Matching module.
}
\label{tab:architecture}
\end{table}

%% file: tables/ldm256.tex
\begin{table}[tbph]
    \setlength\tabcolsep{10.pt}

    \centering   
\begin{tabular}{c||c|c}
    \toprule
    \textbf{CFG scale} & \textbf{FID} $\downarrow$ & \textbf{CLIP} $\uparrow$ \\
    \midrule
    1            & 41.95                     &  0.157  \\
    3            & 17.13                     &  0.214  \\
    5            & 14.03                     &  0.231  \\
    7            & 13.47                     &  0.239  \\
    9            & 13.50                     &  0.243  \\
    \bottomrule
\end{tabular}
\caption{Results of our fine-tuned Latent Diffusion Model on the LAION dataset for different Classifier-free guidance (CFG) scales \cite{ho2022cfg}.}
\vspace{-3mm}
\label{tab:ldm256}
\end{table}